\documentclass[review,times,1p,10pt]{elsarticle}
\usepackage[namelimits]{amsmath}
\usepackage{amssymb, pifont}
\usepackage{multirow}
\usepackage{tabularx}
\usepackage{makecell}
\usepackage{fontawesome}
\usepackage{subfigure}
\usepackage{setspace}
\usepackage{array}
\usepackage[table,xcdraw]{xcolor}
\usepackage[normalem]{ulem}
\useunder{\uline}{\ul}{}
\usepackage{float}
\usepackage[algoruled,lined,boxed,commentsnumbered]{algorithm2e}
\usepackage{enumitem}

\usepackage{hyperref}
\usepackage{caption}
\usepackage{dsfont}
\usepackage{booktabs} 
\usepackage[switch]{lineno}
\usepackage{diagbox}
\usepackage{url}
\usepackage{booktabs}    
\usepackage{xcolor}      
\usepackage{colortbl}    
\definecolor{col}{RGB}{230,230,230} 
\usepackage{threeparttable}
\usepackage{arydshln}   
\usepackage{subcaption} 
\usepackage{mathtools}
\SetAlFnt{\small} 
\SetAlCapSkip{0.5em}
\DontPrintSemicolon 

\begin{document}
	\captionsetup[figure]{labelfont={bf},labelformat={default},labelsep=period,name={Fig.}}
	
	\begin{frontmatter}
		\title{\texorpdfstring{$A^3$-FPN}{A3-FPN}: Asymptotic Content-Aware Pyramid Attention Network for Dense Visual Prediction}
		\author[a,b,c]{Meng'en Qin}
		\ead{mengenching@gmail.com}
		\author[a]{Yu Song}
		\author[a]{Quanling Zhao}
		\author[b]{Xiaodong Yang}
		\author[a]{\texorpdfstring{Yingtao Che\corref{cor1}}{Yingtao Che}}
		\author[a]{\texorpdfstring{Xiaohui Yang\corref{cor1}}{Xiaohui Yang}}
		\address[a]{Henan Engineering Research Center for Artificial Intelligence Theory and Algorithms, Henan University, Kaifeng, China}
		\address[b]{Faculty of Computer Science and Control Engineering, Shenzhen University of Advanced Technology, Shenzhen, China}
		\address[c]{Department of Electrical and Electronic Engineering, The Hong Kong Polytechnic University, Hong Kong, China}
		
		\cortext[cor1]{Corresponding author
		}

		\begin{abstract}
			Learning multi-scale representations is the common strategy to tackle object scale variation in dense prediction tasks. Although existing feature pyramid networks have greatly advanced visual recognition, inherent design defects inhibit them from capturing discriminative features and recognizing small objects. In this work, we propose Asymptotic Content-Aware Pyramid Attention Network ($\boldsymbol{A^3}$-FPN), to augment multi-scale feature representation via the asymptotically disentangled framework and content-aware attention modules. Specifically, $\boldsymbol{A^3}$-FPN employs a horizontally-spread column network that enables asymptotically global feature interaction and disentangles each level from all hierarchical representations. In feature fusion, it collects supplementary content from the adjacent level to generate position-wise offsets and weights for context-aware resampling, and learns deep context reweights to improve intra-category similarity. In feature reassembly, it further strengthens intra-scale discriminative feature learning and reassembles redundant features based on information content and spatial variation of feature maps. Extensive experiments on MS COCO, VisDrone2019-DET and Cityscapes demonstrate that $\boldsymbol{A^3}$-FPN can be easily integrated into state-of-the-art CNN and Transformer-based architectures, yielding remarkable performance gains. Notably, when paired with OneFormer and Swin-L backbone, $\boldsymbol{A^3}$-FPN achieves 49.6 mask AP on MS COCO and 85.6 mIoU on Cityscapes. Codes are available at \url{https://github.com/mason-ching/A3-FPN}.
		\end{abstract}
	\end{frontmatter}
	
	\section{Introduction}
	Dense visual prediction \cite{dense_pr, cednet_pr} is one collection of computer vision tasks that aims to predict the label of every pixel in images. It plays a pivotal role in understanding scenes and is of great importance in autonomous driving and medical imaging, to name a few. With significant breakthroughs in deep learning, a series of promising and leading research is proposed based on Convolutional Neural Networks and Vision Transformers across various dense prediction tasks, including object detection \cite{reppointsv2, small_pr}, instance segmentation \cite{he2017mask, dynamask}, semantic segmentation \cite{fcn_semantic, mask_dino}, etc.
	
\begin{figure}[t]
		\centering
		\includegraphics[scale=0.6]{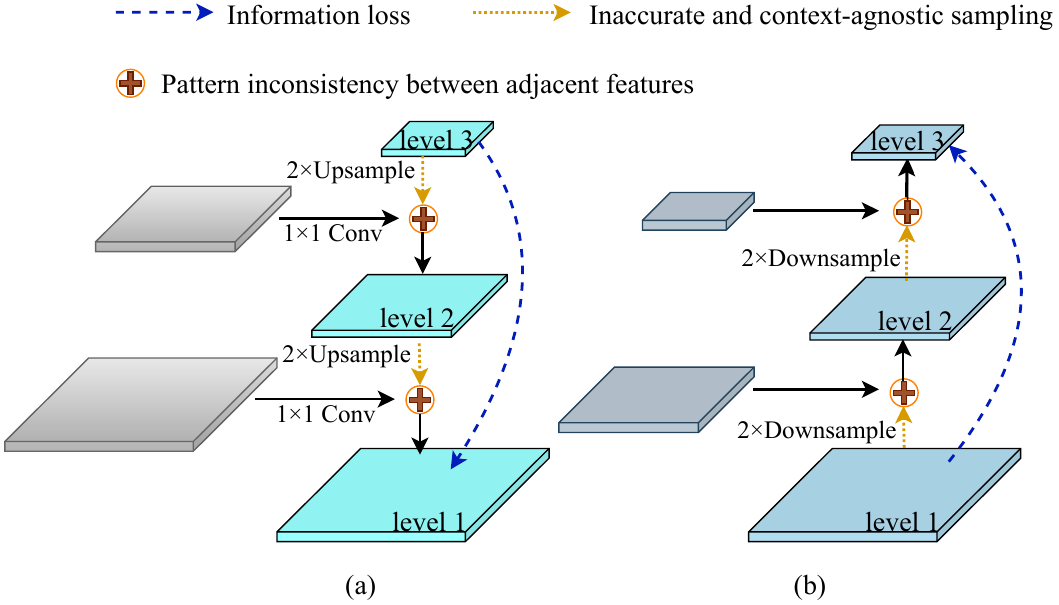}
		\caption{Illustration of FPN \cite{fpn} and PAFPN \cite{liu2018path}. (a) Top-down multi-scale feature fusion path in FPN. (b) Extra bottom-up path aggregation in PAFPN. Both methods have some defects: (1) information loss, (2) context-agnostic sampling, (3) pattern inconsistency.}
		\label{fig:fpn}
	\end{figure}

	Visual prediction tasks need both spatial details for object segmentation or location and semantic information for object classification, which are more likely to reside in different-scale feature maps. Furthermore, objects of different sizes also tend to be recognized on different resolution feature maps. Thus, how to efficiently learn a hierarchy of features at different scales becomes one of the key problems in deep learning methods for visual prediction tasks. Feature Pyramid Network (FPN) \cite{fpn} is the widely used architecture to generate multi-scale features and address the variable size challenge. Specifically, FPN initially employs 1$\times$1 convolution operations to reduce the channel dimension of feature maps extracted from the backbone. Subsequently, it constructs a top-down network to propagate top-level semantic information to lower levels, which consists of two stages: upsample each channel of higher-level feature maps, and directly add the two-level feature maps point by point. Although FPN has substantially improved the performance of deep networks in dense visual tasks, some design defects (e.g. information loss, context-agnostic sampling and pattern inconsistency) inhibit it from further learning more discriminative features, as shown in Fig. \ref{fig:fpn}.
	
	\textbf{Information loss.} FPN adopts a layer-wise top-down pathway to fuse multi-level features, which results in a disadvantageous case: features from a certain level can only sufficiently benefit neighboring-level features, and the interaction between features from non-adjacent levels is weakened. For instance, if features from level 1 intend to access level-3 information, FPN must first fuse the information from level 2 and level 3, and then allow level 1 to indirectly acquire level-3 features by combining the information from level 2. In the top-down propagation, some information in level 3 is more likely to be lost. The above defect still exists in FPN variants like PAFPN \cite{liu2018path} and BiFPN \cite{efficientdet}, which fuse multi-scale features in a layer-by-layer manner.
	
	\textbf{Context-agnostic sampling:} Prior to feature fusion, feature maps are upsampled by interpolation in the top-down branch (FPN) or downsampled through $3\times3$ convolutions in the bottom-up branch (PAFPN). But both the interpolation and strided convolution are context-agnostic and rely on fixed, static sampling patterns based solely on relative sub-pixel neighborhoods. As a result, these operations are prone to introducing inaccurate, redundant, and even erroneous boundary information at the pixel level during subsequent fusion stages. In this situation, objects similar to the background are most probably located inaccurately or even ignored. Several works, such as Dysample \cite{dysample}, CARAFE \cite{carafe++} and FaPN \cite{fapn}, propose sampling-based or kernel-based upsampling operators to enhance feature fusion and mitigate blurred boundary features. Nevertheless, these approaches fail to fully exploit the rich context inherent in the multi-scale hierarchy. Moreover, in addition to upsampling, enhancing downsampling remains unaddressed and is also crucial for accurate and robust object recognition. 
	
	\textbf{Pattern inconsistency:} In the fusion stage of FPN, adjoining features are typically fused via element-wise summation. Because of convolutions and pooling operations between different pyramid levels, there are significant content and pattern inconsistencies between these levels. Element-wise addition fails to address these semantic discrepancies and disregards the underlying relationship learning of different feature representation patterns in the multi-scale fusion, which may lead to intra-category conflict and too many false positive samples in the fused features. AFPN \cite{yang2023afpn} attempts to alleviate cross-scale pattern inconsistency by adaptive spatial fusion, but its performance remains limited due to lacking deep learning of the context relationship of different levels. Most recently, FreqFusion \cite{frequency_aware} introduces a frequency-aware feature fusion method to enhance high-frequency detailed boundary information and reduce semantic inconsistency during upsampling. Despite its improvements, FreqFusion still utilizes a straightforward summation operation and neglects to explicitly model the pattern proportion relationship of different scales. 
	
	Motivated by these issues, we propose Asymptotic Content-Aware Pyramid Attention Network ($A^3$-FPN), to strengthen multi-scale feature representation through the asymptotically disentangled framework and content-aware attention-based feature fusion and reassembly. Compared to other current pyramid networks which inherit FPN's grid-like framework, $A^3$-FPN is distinctive in the following three aspects: (1) Instead of the limited layer-by-layer pathway, it employs a horizontally-spread column-wise network that carries some theoretical advantages to alleviate information loss and help disentangle information of each level from all levels. (2) In feature fusion, it collects supplementary content from adjacent features to produce context-aware offsets and weights for feature resampling, and applies multi-scale context reweighting to model deep pattern relationships and enhance intra-category similarity. (3) In feature reassembly, it further facilitates intra-scale representative content and filters out redundant ones for more accurate and quality locating by reassembling less expressive features based on information density and spatial variation of feature maps.
	
	\begin{figure}[t]
		\centering
		\includegraphics[width=0.9\linewidth]{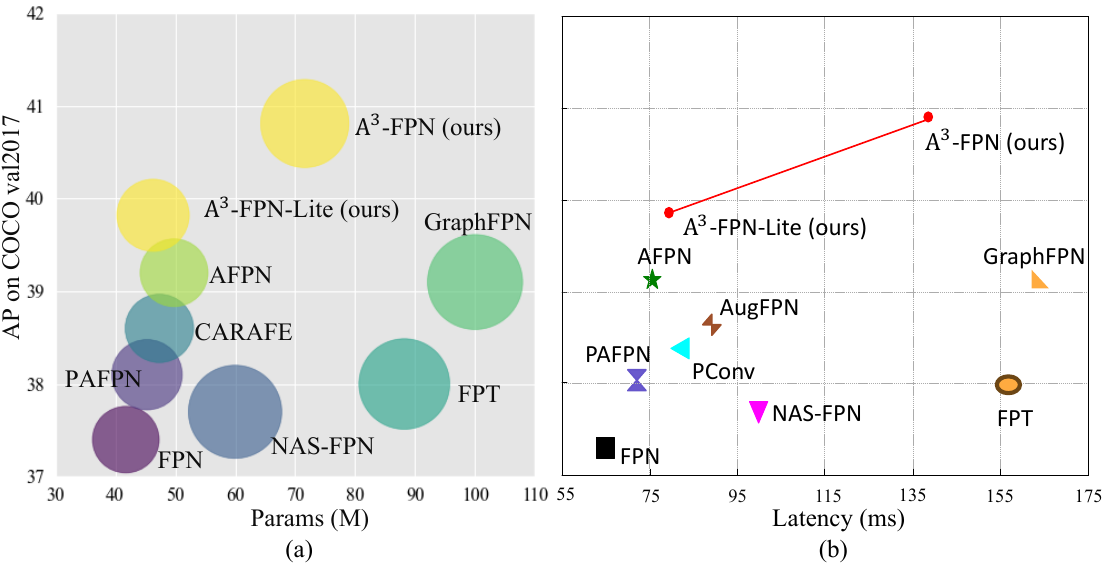}
		\caption{(a) Average precision, parameters, and FLOPs of various feature pyramid networks evaluated on COCO {\tt val2017} \cite{coco}. Bubble area scales with model GFLOPs; (b) Inference latency vs. performance on COCO {\tt val2017} for feature pyramid models. All models are trained for 12 epochs using Faster R-CNN with ResNet-50 as the baseline. Inference latency is measured on a single NVIDIA RTX 4090 GPU.}
		\label{fig:flops_accuracy}
	\end{figure}
	To demonstrate the effectiveness of our model, we implement numerous experiments on MS COCO \cite{coco} and Cityscapes \cite{cityscapes} datasets. Using Faster R-CNN/Mask R-CNN and ResNet-50 as the baseline, $A^3$-FPN achieves a superior performance of 40.9 AP$^{\text{box}}$ and 37.1 AP$^{\text{mask}}$ respectively; the light version, $A^3$-FPN-Lite, attains 39.7 AP$^{\text{box}}$ and 36.3 AP$^{\text{mask}}$ while maintaining fewer parameters and lower latency, as shown in Fig. \ref{fig:flops_accuracy}. In this work, our main contributions are as follows:
	\begin{itemize}
		\item 
		We present an in-depth analysis of existing multi-scale feature fusion frameworks and operations widely used in dense visual prediction tasks. Specifically, we identify three critical issues: information loss, context-agnostic sampling and pattern inconsistency, which hinder expressive hierarchical feature learning, and result in intra-category inconsistency and object displacement.
		
		\item 
		$A^3$-FPN, an asymptotic content-aware pyramid attention network, is proposed to tackle these issues. The asymptotically disentangled framework can effectively alleviate information loss and gradually learn required information in the asymptotically global feature interaction. The content-aware attention-based fusion modules are designed to mitigate intra-category content inconsistency and boundary displacement for medium-to-large objects, while facilitating discriminative content learning for small instances.
		
		\item 
		Qualitative and quantitative results show that $A^3$-FPN and $A^3$-FPN-Lite can be easily integrated into SOTA dense prediction CNNs and Vision Transformers, leading to considerable improvements. It is worth mentioning that $A^3$-FPN with OneFormer as the architecture and Swin-L as the backbone accomplishes a new record of 49.6 AP$^{\text{mask}}$ on COCO val2017 and 85.6 mIoU on Cityscapes.
	\end{itemize}
	\section{Related Work}
	\subsection{Dense Visual Prediction}
	Dense prediction tasks \cite{dense_pr, cednet_pr} encompass a range of challenges, including object detection \cite{reppointsv2, small_pr}, instance segmentation \cite{he2017mask, dynamask}, semantic segmentation \cite{fcn_semantic, mask_dino}, etc. Advancements in dense visual prediction have been primarily driven by several seminal deep learning architectures. For semantic segmentation, fully convolutional network (FCN) \cite{fcn_semantic} marks a turning point and leads to the development of fundamental frameworks such as PSPNet \cite{pspnet}, PSANet \cite{psanet}, UperNet \cite{upernet}, PointRend \cite{pointrend} and SegNext \cite{segnext}. Transformer-based segmentors like SETR \cite{setr} and SegFormer \cite{segformer} utilize the self-attention mechanism to capture long-range contextual dependencies, achieving state-of-the-art performance on semantic segmentation benchmarks. In object detection, R-CNN series \cite{faster_rcnn, cai2018cascade} and YOLO series \cite{yolov10} have become the dominant CNN-based methods, achieving a great balance between model performance and inference speed. Additionally, detection transformers (DETRs) make their debut in DETR \cite{detr}, eliminating traditional hand-designed components and NMS. As the first real-time detection transformer, RT-DETR \cite{rtdetrv1} designs an efficient hybrid encoder to process multi-scale features and uncertainty-minimal query selection to provide high-quality initial queries to the decoder, therefore improving detection accuracy. As for instance segmentation, Mask R-CNN \cite{he2017mask} and its variants \cite{cai2018cascade, dynamask} have yielded promising pixel-level results by adding extra mask prediction branches to Faster R-CNN. Most recently, some models, such as Mask2Former \cite{mask2former}, Mask DINO \cite{mask_dino} and OneFormer \cite{oneformer}, have advanced the field by unifying segmentation and detection tasks within a single framework.
	\subsection{Multi-scale Feature Representation}
	Typically, features at different levels encode positional information corresponding to objects of different sizes. Small feature maps carry abstract features and position information of large objects, while large feature maps capture low-dimensional textures and position details of small objects \cite{goldyolo}. FPN \cite{fpn} leverages the feature diversity and fuses multi-scale features by cross-scale connections, thereby increasing detection accuracy for objects of varied sizes. Building on FPN, PANet \cite{liu2018path} integrates a bottom-up pathway for more comprehensive information fusion across different levels. EfficientDet \cite{efficientdet} introduces BiFPN, a novel and repeatable module that enhances the efficiency of information fusion. FaPN \cite{fapn} improves FPN for dense visual predictions by aligning upsampled feature maps. AFPN \cite{yang2023afpn} promotes feature interaction across non-adjacent levels but ignores information redundancy and object displacement in feature fusion. NAS-FPN \cite{nas-fpn} leverages Neural Architecture Search (NAS) to automatically discover the optimal feature pyramid framework in a data-driven manner. SFI network \cite{enhancing_sfi} leverages frequency-domain information to improve fused FPN features and LR-FPN \cite{lr-fpn} introduces SPIEM and CIM to enable finer multi-scale interaction and to reinforce object-region representations. GraphFPN \cite{zhao2021graphfpn} adopts a graph neural network to enable non-adjacent feature interaction and information propagation across the pyramid, but its graph-based modeling leads to a substantial increase in parameters and computational overhead. Gold-YOLO \cite{goldyolo} enables straight interaction across different levels and achieves better fusion through the Gather-and-Distribute mechanism. 
	$A^2$-FPN \cite{a2-fpn} proposes MGC, GACARAFE, and GACAP to address inaccurate sampling and semantic inconsistency, but it ignores information transmission loss caused by the overall fusion framework. We propose the asymptotically disentangled framework to explicitly alleviate such framework-induced information loss. Moreover, $A^2$-FPN’s kernel-based sampling is spatially fixed and thus less effective for objects of varied shapes. 
	TFPN \cite{TFPN} designs feature reference, calibration, and feedback modules with the same motivations as $A^2$-FPN. Although TFPN’s upsampling and fusion operations are similar to ours, there are some key differences: (1) our method learns much richer multi-level context via the offset generator that guides the resampler; (2) TFPN assumes shared offset maps for all channels during calibration, but we introduce the grouping strategy so that each group of channels corresponds to a group of offset maps, increasing model diversity and robustness.
	\begin{figure}[t]
		\centering
		\includegraphics[width=0.85\linewidth]{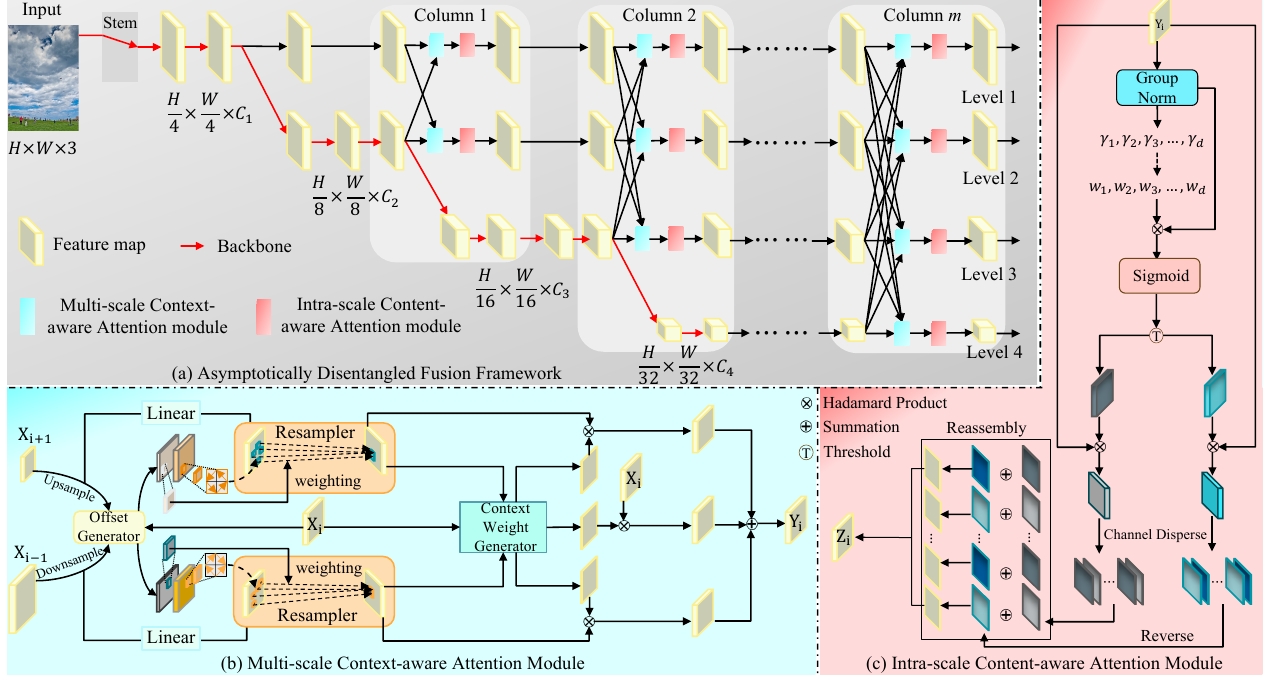}
		\caption{Overall architecture of $A^3$-FPN. (a) The bottom-up asymptotically disentangled fusion framework consisting of $m$ columns; (b) Multi-scale Context-aware Attention module for feature fusion; (c) Intra-scale Content-aware Attention module for feature reassembly.}
		\label{fig:a3fpn}
	\end{figure}
	
	\section{Method}
	Fig. \ref{fig:a3fpn} presents the general architecture of $A^3$-FPN, which consists of three essential components: asymptotically disentangled framework, multi-scale context-aware attention module for feature fusion (MCAtten), and intra-scale content-aware attention module for feature reassembly (ICAtten). We also develop $A^3$-FPN-Lite, a more efficient and lightweight version, which only differs from $A^3$-FPN in some hyperparameter settings (refer to Appendix D). For convenient discussion, we use $X_i$ to denote the i-th level input features in each column. $Y_i$ is the i-th level fused features from MCAtten, and $Z_i$ is the i-th level reassembled features from ICAtten. The whole algorithm procedure of $A^3$-FPN is summarized in Appendix A.
	\begin{figure}[t]
		\centering
		\includegraphics[width=0.9\linewidth]{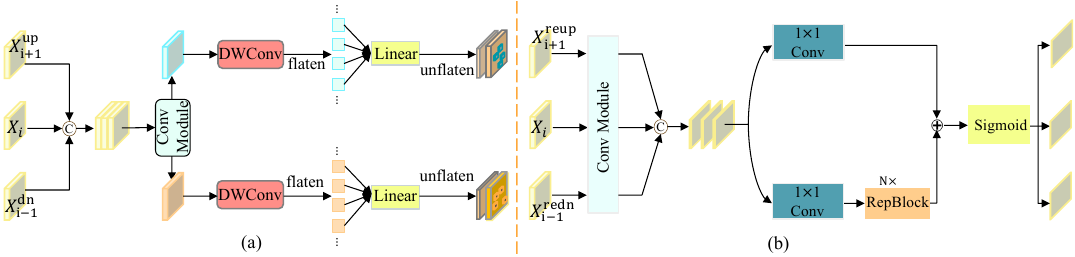}
		\caption{Offset generator and context weight generator in multi-scale context-aware attention module. (a) Offset generator gathers context information to produce position-wise coordinate offset maps and sampling weight maps for the subsequent Resampler; (b) Context weight generator learns the relationship among different representation patterns and assigns the corresponding context weight to different-level features.}
		\label{fig:generator}
	\end{figure}
	\subsection{Asymptotically Disentangled Framework}
	As shown in Fig. \ref{fig:framework_comparsion}, existing pyramid networks can be categorized into layer-wise or global feature fusion frameworks. Layer-wise frameworks (e.g., FPN \cite{fpn}, PANet \cite{liu2018path} and BiFPN \cite{efficientdet}) only utilize local information in each feature fusion and facilitate information flow in a layer-by-layer manner, which hinders the feature utilization among non-adjacent levels. Global frameworks, such as Deformable Attention \cite{deformable_detr} and Gold-YOLO \cite{goldyolo}, make sure that any level can directly acquire information from all other levels. While the global framework can enhance multi-scale feature interaction, it overlooks the significant pattern and content gaps from non-adjacent levels, especially for the bottom and topmost features. Inspired by High Resolution Networks \cite{HRNEtforvisual}, we propose an asymptotic column-spread framework in $A^3$-FPN to represent multi-scale features. It initiates the feature fusion by combining adjacent-level features and progressively disentangles every level from all levels through column-wise information interaction and horizontal information spread. The proposed framework carries some great theoretical advantages over others, including (1) sufficient and asymptotically direct interaction among features from all levels, alleviating information loss in the transmission path; (2) helping learn the superior transformation function to disentangle information needed for dense visual predictions at each level. These properties are guaranteed by the following lemmas: 
	\begin{figure}[t]
		\centering
		\includegraphics[width=0.9\linewidth]{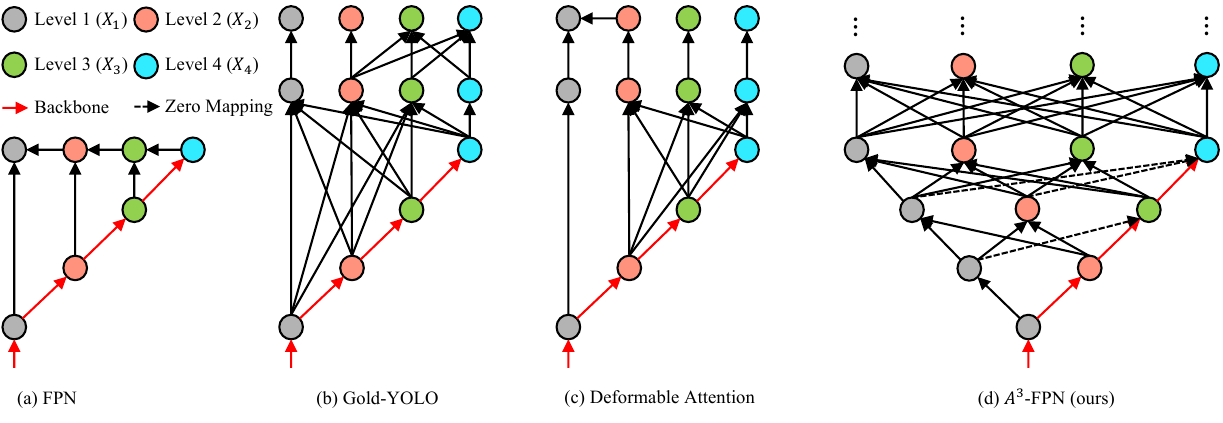}
		\caption{Comparison of different multi-scale designs, including (a) FPN \cite{fpn} (layer-wise framework), (b) Gold-YOLO \cite{goldyolo} (global convolutional framework), (c) Deformable Attention \cite{deformable_detr} (global attention framework), (d) $A^3$-FPN (asymptotically disentangled framework).}
		\label{fig:framework_comparsion}
	\end{figure}
	
	\textbf{Lemma 1}. Let $X_i$ denote the random variable corresponding to the input feature at level $i$. Suppose a framework induces a propagation path from $X_i$ to $X_j$ (i.e. $X_i\rightarrow X_{i+1}\rightarrow\cdots\rightarrow X_j$). If each step $X_t\rightarrow X_{t+1}$ satisfies a strong data processing inequality with contraction coefficient $\eta_t\in[0,\ 1]$, the retained information from $X_i$ to $X_j$ is bounded by 
	\[
	I(X_i;\ X_j)\le(\prod_{t=1}^{j-i}\eta_t)\ H(X_i),
	\]
	where $H(X_i)<\infty$ is the entropy of $X_i$, $I(;)$ denotes the mutual information.
	
	\textbf{\textit{Proof}}. In fact, information propagation between the input feature $X_i$ and $X_j$ can be represented as a directed Markov chain of intermediate random variables: $X_i\rightarrow X_{i+1}\rightarrow\cdots\rightarrow X_{j-i}\rightarrow X_j$. The chain is the actual directed sequence of stochastic mappings along the shortest path. Apply the strong data processing inequality successively down the chain, then we can get 
	\begin{equation}
		\begin{matrix}
			I\left(X_i;X_{i+1}\right)\le\eta_0I\left(X_i;X_i\right)=H\left(X_i\right),
			\\I\left(X_i;X_{i+2}\right)\le\eta_1I\left(X_i;X_{i+1}\right)\le\eta_1\eta_0H\left(X_i\right),
			\\\vdots
			\\I\left(X_i;X_j\right)\le\left(\prod_{t=1}^{j-i}\eta_t\right)H\left(X_i\right).
		\end{matrix}
	\end{equation}
	This completes the proof of Lemma 1. We note that if step t is deterministic or invertible, the corresponding $\eta_t$ equals 1. Real network operations (downsampling, bottlenecks, activation nonlinearities, training stochasticity, etc.) often act as non-invertible/noisy steps for which $\eta_t<1$ is plausible. \hfill $\blacksquare$
	
	\textbf{Lemma 2}. Given a compact set $\mathcal{K}$, $\mathcal{F}^\ast:\ \mathcal{K}\rightarrow\ \mathcal{K}^d$ is continuous and admits a composition of finitely many local continuous maps arranged on a directed acyclic graph (DAG) $\mathcal{G}$. If each local map in the DAG is $L_v$-Lipschitz on its compact domain, for every $\varepsilon>0$, there exists a finite column count $m$ and a parametric continuous mapping for each column such that the asymptotic composition $\mathcal{F}_m$ produced by these columns satisfies 
	\[
	\sup_{x \in K} \, \| \mathcal{F}_m(x) - \mathcal{F}^\ast(x) \| \le \varepsilon.
	\]
	In other words, an asymptotic column-wise network can uniformly approximate $\mathcal{F}^\ast$ on $\mathcal{K}$ arbitrarily well.
	
	\textbf{\textit{Proof}}. $\mathcal{G}=(\mathcal{V},\ \mathcal{E})$ is a finite directed acyclic graph, where $\mathcal{V}_{in}$ are the inputs and root (output) nodes provide $\mathcal{F}^\ast(x)$. For input nodes $\nu_i^{in}\in\mathcal{V}_{in}$, set $y_{\nu_i^{in}}(x)=\hat{y}_{\nu_i^{in}}(x)=x_i$ so their node errors are zero. For each non-input node $\nu \in \mathcal{V} \setminus \mathcal{V}_{in},$ there is a continuous local mapping:
		$\phi_\nu:\ \prod_{u\in p a(\nu)}{\mathcal{Y}_u\rightarrow}\ \mathcal{Y}_\nu,$
	where $pa(\nu)$ are the parents of $\nu$ in $\mathcal{G}$ and $\mathcal{Y}_u$ is the output space of node $u$. The value at node $\nu$ is $\mathcal{Y}_\nu=\phi_\nu(\left\{\mathcal{Y}_u:\ u\in p a(\nu)\right\})$.
	
	Because $\mathcal{F}^\ast$ is represented by $\mathcal{G}$ and $\mathcal{G}$ is finite, index the non-input nodes in a topological order $\nu_1,\ \nu_2,\ \cdots,\ \nu_n$ so that all parents of $\nu_i$ have indices $<i$. For each $\nu_i$, the mapping $\phi_{\nu_i}$ is continuous and $L_{\nu_i}$-Lipschitz on a compact domain $D_i$. 
	Fix any collection of approximations ${\hat{\phi}}_{\nu_i}$. Define the approximated node values ${\hat{\mathcal{Y}}}_\nu$ by replacing $\phi_\nu$ with ${\hat{\phi}}_\nu$ and evaluate the DAG in the same topological order. For each node $\nu_i$, let the uniform per-node approximation error be
		$\varepsilon_i=\sup_{z \in D_i} \, \|\hat{\phi}_{\nu_i}(z)-\phi_{\nu_i}(z)\|.
		$
	For each node $\nu_i$, we can define the maximal path-amplification factor $W_i$ as 
	\begin{equation}
		W_i=\max_{\pi\in\mathrm{\Pi}(i \rightarrow root)} \prod_{u\in\pi} L_u,
	\end{equation}
	where $\mathrm{\Pi}(i\rightarrow root)$ is the set of directed paths in the DAG from node $\nu_i$ to an output/root node, and the product runs over nodes on the path after $\nu_i$. If there is no path from $\nu_i$ to the root, set $W_i=0$. Then the total uniform error at the root induced by replacing every $\phi_{\nu_i}$ with ${\hat{\phi}}_{\nu_i}$ satisfies 
	\begin{equation}
		\label{eq:sum_w}
		\sup_{x \in K} \, \| \mathcal{F}_m(x) - \mathcal{F}^\ast(x) \| \le \sum_{i=1}^{n}W_{i}\varepsilon_{i}.
	\end{equation}
	To guarantee the right-hand side of Equation \ref{eq:sum_w} $\le\varepsilon$, it suffices to choose any positive $\varepsilon_i$ satisfying 
		$\sum_{i=1}^{n}{W_i\varepsilon_i} \le \varepsilon$.
	A simple explicit choice is to set, for example,
	\begin{equation}
		\label{eq:varepsilon_i}
		\varepsilon_i =
		\begin{cases}
			\dfrac{\varepsilon}{n W_i}, & \text{when } W_i > 0, \\[8pt]
			0, & \text{when } W_i = 0,
		\end{cases}
	\end{equation}
	then
	\begin{equation}
		\sum_{i=1}^{n} W_i\varepsilon_i=\sum_{i:W_i>0}^{n} W_i\cdot\frac{\varepsilon}{nW_i}=\varepsilon\cdot\frac{\left|\left\{i:W_i>0\right\}\right|}{n}\le\varepsilon.
	\end{equation}
	Each $\phi_{\nu_i}$ is continuous on compact domain $D_i$. By the universal approximation theorem \cite{sigmoid_approximation} and its recent advances \cite{mlp_width}, for any $\delta>0$, there exists a finite-width finite-depth parametric neural network (MLP with a standard non-polynomial activation like ReLU or sigmoid, possibly arranged into a conv block) that uniformly approximates $\phi_{\nu_i}$ within $\delta$ on $D_i$. Therefore, for each $i$ we can choose a finite ${\hat{\phi}}_{\nu_i}$ such that
		$\sup_{z\in D_i} \|\hat{\phi}_{v_i}(z)-\phi_{v_i}(z)\| \le \varepsilon_i$
	using $\varepsilon_i$ chosen in Equation \ref{eq:varepsilon_i}.
	
	Map nodes to columns according to a topological layering of $\mathcal{G}$: every column $C^{(k)}$ implements the collection of approximators $\left\{{\hat{\phi}}_\nu\right\}$ for those nodes assigned to column $k$. Because the DAG and per-node approximators are finite, the number of columns $m$ is finite, and each column is a finite parametric continuous map. The composition of columns $\mathcal{F}_m=C^{(m)}\circ\cdots\circ C^{(1)}$ is precisely the global approximator of $\mathcal{F}^\ast$. 
	By construction, the per-node uniform errors are the $\varepsilon_i$ chosen in Equation \ref{eq:varepsilon_i}, so by Equation \ref{eq:sum_w} the uniform approximation error satisfies
	\begin{equation}
		\sup_{x \in K} \, \| \mathcal{F}_m(x) - \mathcal{F}^\ast(x) \| \le \sum_{i=1}^{n}W_{i}\varepsilon_{i} \le \varepsilon.
	\end{equation}
	Hence, the finite-column finite-width asymptotically disentangled framework approximates $\mathcal{F}^\ast$ within $\varepsilon$ uniformly on $\mathcal{K}$. This completes the proof of Lemma 2. \hfill $\blacksquare$
	
	Lemma 1 shows that, under realistic non-invertible mappings, the shorter-path design has a higher transmission upper bound, permitting more information to be retained among multi-level features. This explains why frameworks that create direct paths between two-level features (e.g., global fusion and asymptotically disentangled fusion) are less likely to suffer cumulative information loss than those requiring many sequential processing hops to transfer the same information (layer-by-layer fusion). For empirical validation, we provide a detailed performance analysis of different fusion frameworks with the same fusion operations in Table \ref{tab:ablation_framework}, which further illustrates the superiority of the proposed framework.
	
	In Fig. \ref{fig:framework_comparsion} (d), if considering each column in $A^3$-FPN as one layer of MLP, the proposed framework is indeed an MLP-DAG-style network. In fact, the overall objective of networks can be interpreted as learning an optimal network or function $\mathcal{F}^\ast$. Given an image $x$, the network maps $x$ to $\mathcal{F}^\ast(x)$ such that the output representation forms a hierarchy of features that are well aligned with downstream tasks. The hierarchical features can acquire precisely the information required at that level by $\mathcal{F}^\ast$. Since $\mathcal{F}^\ast$ is realized by a feed-forward architecture, information flows strictly forward without cycles, which naturally satisfies the definition of a directed acyclic graph (DAG): a decomposition as a composition of finitely many local continuous maps. Lemma 2 ensures that, compared with other fusion frameworks, the asymptotically disentangled design can approximate the global optimum of $\mathcal{F}^\ast$ more effectively with finite column depth and width. Considering practical constraints such as runtime and computational cost, we fix the number of columns to three, and set the max width of columns to four.
	
	Moreover, there are two design paradigms for $A^3$-FPN: top-down or bottom-up asymptotic framework (see top-down asymptotic framework in Appendix B). In section \ref{ablation}, we will experimentally prove that the former is more appropriate for position-relevant tasks (object detection and instance segmentation), whereas the latter is more suited to semantic segmentation. 
	\subsection{Multi-scale Context-aware Attention for Feature Fusion}
	The conventional feature fusion suffers from two issues that detrimentally impact dense visual prediction, namely intra-category inconsistency and object displacement, which are mainly caused by inaccurate sampling and pattern-agnostic fusion operations. As shown in Fig. \ref{fig:a3fpn} (b), we propose MCAtten to align object-level features and strengthen intra-category similarity across different levels. MCAtten is comprised of two steps: context attention-guided feature resampling and feature reweighting, each of which will be discussed in detail below. 
	
	\textbf{Position-wise offset generator.} Given i-th level features $X_i\in\mathbb{R}^{c_i\times h_i\times w_i}$ of (m-j)-th column, where $i \leq \textbf{min}=\min(m-j+1, n)$, $j \in \{0, 1, \ldots, m-1\}$, $n$ is the number of levels, $c_i$ denotes the channel dimension and $h_i$, $w_i$ are the spatial size of $X_i$, we upsample $\left\{X_{i+1}, \cdots, X_{\textbf{min}}\right\}$ by $1\times1$ convolutions and bilinear interpolations (nearest neighbor) and downsample $\left\{X_1, \cdots, X_{i-1}\right\}$ by strided convolutions with different strides and kernels. Now we attain the coarsely sampled features $\left\{X_1^{up}, \cdots, X_{i-1}^{up}, X_{i+1}^{dn}, \cdots, X_{\textbf{min}}^{dn} \right\}$, each of which has the same shape as $X_i$. Subsequently, $\left\{X_1^{up}, \cdots, X_i, \cdots, X_{\textbf{min}}^{dn} \right\}$ are fed into the offset generator to learn the position and context semantic information of previous coarse sampling points. The offset generator shown in Fig. \ref{fig:generator} (a) begins the process by concatenating all inputs and then puts the concatenation into the convolution module to produce context information with $(\textbf{min}-1)c_i$ channels. The context content is evenly split into $(\textbf{min}-1)$ parts, each of which will go through an individual offset generator branch to attain the specific offsets and weights for every sampled level. In the offset generator branch, we utilize a depthwise convolution and linear projection layer to process the context information, finally yielding fine-grained and position-wise offsets $X_{offset}\in\mathbb{R}^{K^2\times 3 \times h_i\times w_i}$ for context-aware feature resampling. However, it may impair model generalization and diversity that all $c_i$ channels share the same offset maps. Therefore, we divide the $c_i$ channels into several groups, for each of which there is a learned $X_{offset}$. Eventually, the offset $X_{offset}^G$ belongs to $\mathbb{R}^{G \times K^2\times 3 \times h_i\times w_i}$, where $G$ means the number of groups, $K^2$ denotes $K \times K$ resampling points on sampled feature maps. Additionally, in $X_{offset}^G$ every resampling point $(x, y)$ corresponds to a pair of coordinate offsets $(\Delta x, \Delta y)$ and coordinate attention weight $\Delta m$, which indicates the significance of the sampling point.
	
\begin{figure}[t]
		\centering
		\includegraphics[width=1.0\linewidth]{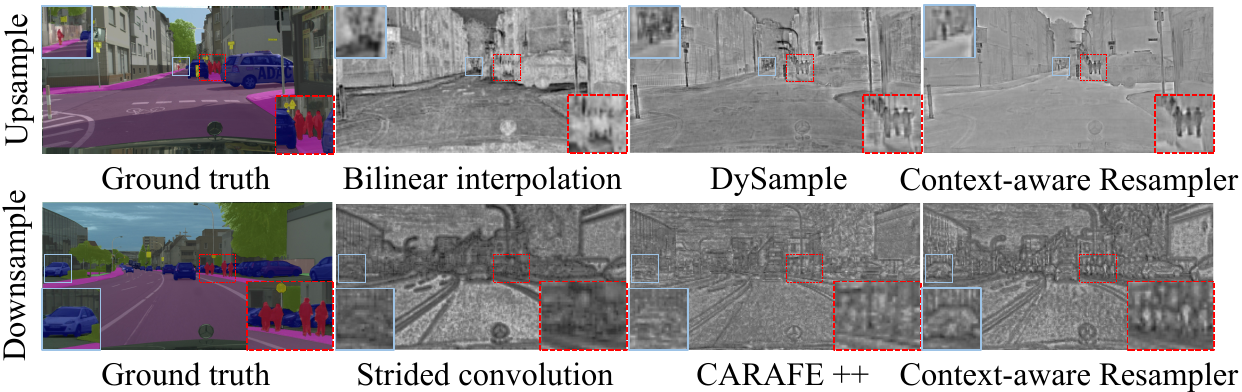}
		\caption{In first row, $H/8 \times W/8$ feature maps are upsampled to $H/4 \times W/4$ by bilinesar interpolation, DySample \cite{dysample} and Cotext-aware Resampler (ours); In second row, we downsample $H/4 \times W/4$ feature maps to $H/8 \times W/8$ through strided convolution, CARAFE++ \cite{carafe++} and Cotext-aware Resampler (ours).}
		\label{fig:sample_comparison}
	\end{figure}

	\textbf{Context-aware feature resampler.} After acquiring the offset maps, we exploit deformable convolutions \cite{dcnv4} to resample coarsely sampled feature maps, followed by GELU activation and Layer Normalization. Here, we briefly review the deformable convolution and then explain why it can function as the core operation of Resampler. For a sampled feature map $X^s\in\mathbb{R}^{c_i\times h_i\times w_i}$, the output feature at the position $X[(x, y)]$ after a vanilla convolution with the $K \times K$ kernel can be obtained by:
	\begin{equation}
		X[(x, y)] = \sum_{n=1}^{K^2} w_n \cdot X^s[(x, y) + p_n],
		\label{eq:conv}
	\end{equation}
	where \((x, y) \in \left\{(0, 0), (0, 1), \ldots, (h_i - 1, w_i - 1)\right\}\), $K^2$ is the number of sample points each time, \(w_n\) and \(p_n \in \left\{\left(-\left\lfloor \frac{K}{2} \right\rfloor, -\left\lfloor \frac{K}{2} \right\rfloor\right), \left(-\left\lfloor \frac{K}{2} \right\rfloor, -\left\lfloor \frac{K}{2} \right\rfloor+1\right), \ldots, \left(\left\lfloor \frac{K}{2} \right\rfloor, \left\lfloor \frac{K}{2}\right\rfloor\right)\right\} \) stands for the convolutional weight and pre-defined offset for the n-th sample location, respectively. In addition to the fixed offsets, deformable convolutions attempt to learn extra coordinate offsets $(\Delta x, \Delta y)$. When applying deformable convolutions to $X^s$, Equation \ref{eq:conv} can be reformulated as:
	\begin{equation}
		X^{rs} = \sum_{n=1}^{K^2} w_n \cdot X^s[(x, y) + p_n + (\Delta x_n, \Delta y_n)] \cdot \Delta m_n,
		\label{eq:deformable}
	\end{equation}
	where $(\Delta x_n, \Delta y_n)$ and $\Delta m_n$ are the learnable coordinate offsets and attention weight for the n-th location. Furthermore, considering dividing the channels into G groups, the final resampled features are calculated by:
	\begin{equation}
		X^{rs} = \bigcup_{g=1}^{G} \sum_{n=1}^{K^2} w_{g} \cdot X^s_g[(x, y) + p_n + (\Delta x_{gn}, \Delta y_{gn})] \cdot \Delta m_{gn},
		\label{eq:group_deformable}
	\end{equation} 
	where $\bigcup$ is the concatenation of feature maps, \(G\) is the group number and \(w_g \in \mathbb{R}^{\frac{c_i}{G}}\) represents the location-irrelevant projection weights of the g-th group. 
	Because offset maps are produced by the context-aware offset generator, Equation \ref{eq:deformable} and \ref{eq:group_deformable} can resample the coarse feature maps based on learned context content, correcting object displacement and boundary redundancy. In addition, we compare different upsampling and downsampling methods with the context-aware feature resampler, which is shown in Fig. \ref{fig:sample_comparison}. Our approach significantly restores representative object features in the sampling process, while reducing the misplaced boundary pixels.
	
	\textbf{Context weight generator.} In this section, we need to rethink how to efficiently fuse multi-scale features $\left\{X_1^{rs}, \cdots, X_i, \cdots, X_{\textbf{min}}^{rs} \right\}$. Features from different levels embody different content representation patterns. Some patterns contain more abstract features that are quite essential for recognizing the existence of objects, while others carry more detailed content that is beneficial to understanding the object boundary. Although Resampler mitigates pixel-level feature disharmony and error, direct feature fusion by element-wise summation still exposes relatively low intra-category similarity \cite{frequency_aware}, and also hinders relationship attention learning among different feature patterns, thus increasing the risk of false classification and underdetection. The proposed context weight generator highlights context content modeling and pattern relationship learning in the fusion stage, aiming to reduce the intra-category inconsistency and background interference. In Fig. \ref{fig:generator} (b), the generator first squeezes $\left\{X_1^{rs}, \cdots, X_i, \cdots, X_{\textbf{min}}^{rs} \right\}$ with $\textbf{min}$ convolution modules to avoid drastic channel reduction in the following operations. And then, we feed the concatenated features to two branch paths, where the lower pathway applies N RepBlocks composed of RepConv to extract abstract attention content and the upper branch with a cheap $1\times1$ convolution layer attains the shallow context information as a supplement to the lower branch. Afterward, the two-path outputs are fused by element-wise addition and projected by the Sigmoid function, generating the context attention weight maps $W_i \in \mathbb{R}^{\textbf{min}\times h_i\times w_i}$. Each channel in the context weights represents the pattern proportion relationship of the corresponding level, and the i-th level fused features $Y_i$ are finally calculated by:
	\begin{equation}
		\begin{aligned}
			Y_i=(\sum_{n=1, n \neq i}^{\textbf{min}}{W_i^n}\bigotimes X^{rs}_{n}) + W_i^i \bigotimes X_{i},
		\end{aligned}
	\end{equation}
	where $\bigotimes$ implies Hadamard product, $W_i^n \in \mathbb{R}^{h_i\times w_i}$ refers to the aggregation context weight map of the i-th level and n-th channel in $(m-j)$-th column.
	\begin{figure}[t]
		\centering
		\includegraphics[width=0.9\linewidth]{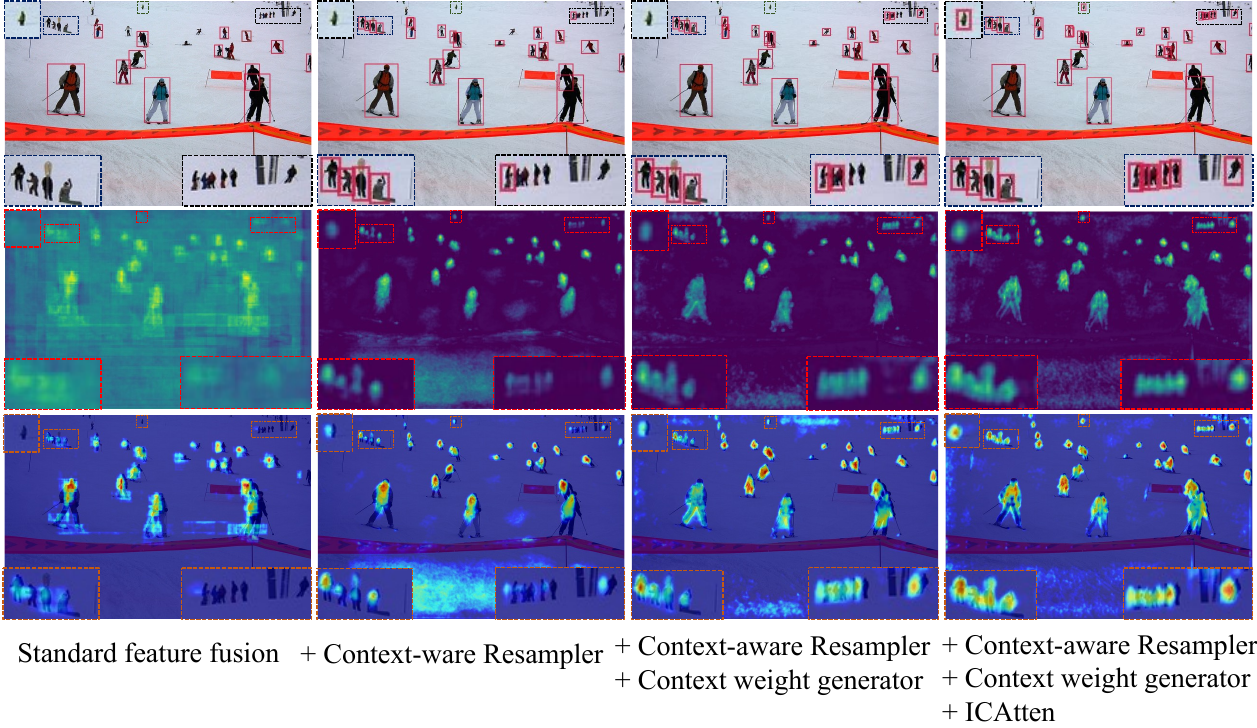}
		\caption{Visualization of detection results, the corresponding feature maps and heatmaps. Resampler refines coarsely sampled features and diminishes object displacement. The context weight generator learns the significance relationship of different feature patterns, decreasing the misclassifications and missed detections. ICAtten further enhances the discriminative features and alleviates complex background interference.}
		\label{fig:ablation_module}
	\end{figure}
	\subsection{Intra-scale Content-aware Attention for Feature Reassembly}
	As is shown in Fig.\ref{fig:ablation_module}, while we notably improve the detection performance with the resampler and context weight generator, some tiny and occluded objects are still located inaccurately or underdetected. In fact, models confront difficulty separating tiny objects from complex and cluttered backgrounds since they take up fewer pixels. Moreover, some fused object features are not discriminative and expressive enough to be recognized by the model, also resulting in misclassifications and missed detections. Motivated by this, we design ICAtten to further suppress the redundant background information by the spatial content variation of feature maps and facilitate intra-scale target-relevant content attention learning by channel reassembly and feature reuse. To fulfill this, we first need to separate the expressive features from less expressive ones. The latter can be regarded as trivia and play a role as a complement to the former. Considering the model's FLOPs and parameters, we use the learnable parameter scaling factors in Group Normalization \cite{gn} to evaluate the information density of feature maps. For fused features $Y_i\in\mathbb{R}^{c_i\times h_i\times w_i}$, we initially subtract the mean and divide $Y_i$ by the standard deviation as follows:
	\begin{equation}
		\begin{aligned}
			Y_i^{\mathrm{std}}=GN(Y_i)=\alpha\frac{Y_i-\mu}{\sqrt{\sigma^2}}+\beta
		\end{aligned}
	\end{equation}
	where $\sigma$ and $\mu$ are the standard deviation and mean of $Y_i$, $\alpha$ and $\beta$ are learnable transformation coefficients. We apply the trainable parameters $\alpha\in R^{c_i}$ in GN to assess the variance of pixel values across batches and channels. Greater variation represents richer spatial and semantic content density. The standardized weights $\omega_i\in R^{c_i}$ are attained by Equation~\ref{omega}, indicating the informativeness of features.
	\begin{equation}    \label{omega}
		\begin{aligned}
			\omega_i=\left\{\frac{\alpha_i}{\sum_{j=1}^{c_i} \alpha_j}, i=1,2, \cdots,c_i\right\}
		\end{aligned}
	\end{equation}
	We reweight the feature values by $\omega_i$ and subsequently transform them to the range $(0,1)$ using the Sigmoid function to calculate the reweights. Afterward, we designate reweights exceeding the threshold as 1 to generate the informative attention weights $\omega_i^1$, whereas reweights falling below the threshold are set as 0 to generate the non-informative weights $\omega_i^2$ (threshold is set to 0.5 in the experiments). The computation of ${\omega_i^1,\ \omega_i^2}$ can be formulated as:
	\begin{equation}
		\begin{aligned}
			\omega_i^1 &= 
			\begin{cases}
				\text{Sigmoid}(\omega_i), & \text{Sigmoid}(\omega_i) \leq \text{threshold}; \\
				1, & \text{Sigmoid}(\omega_i) > \text{threshold},
			\end{cases} \\  
			\omega_i^2 &= 
			\begin{cases} 
				\text{Sigmoid}(\omega_i), & \text{Sigmoid}(\omega_i) \geq \text{threshold}; \\
				0, & \text{Sigmoid}(\omega_i) < \text{threshold}.
			\end{cases}
		\end{aligned}
	\end{equation}
	After that, the fused features $Y_i$ are multiplied by $\omega_i^1$ and $\omega_i^2$, respectively, producing two resulting weighted features: the informative ones $Z_1$ and less informative ones $Z_2$. So far, we have condensed the irrelevant background features and separated the fused features into two parts: $Z_1$ has expressive and informative spatial and semantic information, while $Z_2$ has less information, which can be viewed as trivial details. Subsequently, we disperse $Z_1$ and $Z_2$ along the channel dimension and make each channel of $Z_1$ and its reverse-matched channel of $Z_2$ add together, promoting the information flow efficiency across channels and strengthening the representative feature expression. Finally, we concatenate all reassembled channels to form $Z_i$.
	\section{Experiments}
	\begin{table}[t]
		\centering
		\caption{Performance of Faster RCNN \cite{faster_rcnn} with different feature pyramid networks on MS COCO {\tt val2017}. $^{\dag}$ indicates model performance referred from other works.}
		\renewcommand{\arraystretch}{1.0}
		\label{tab:comparison_fpn}
		\resizebox{0.95\linewidth}{!}{
			\begin{tabular}{l|c|ccc|ccc|ccc}
				\toprule
				Method& Backbone & AP & AP$_{50}$ & AP$_{75}$ & AP$_S$ & AP$_M$ & AP$_L$ & Params(M) & FLOPs(G) \\
				\midrule
				FPN \cite{fpn} & R50 & 37.4 & 57.3 & 40.3 & 18.4 & 41.7 & 52.7 & 41.7 & 187 \\
				PConv$^{\dag}$ \cite{pconv} & R50 & 38.5 & 59.9 & 41.4 & - & - & - & 42.8 & - \\
				FreqFusion \cite{frequency_aware} & R50 & 39.4 & 60.9 & 42.7 & 23.0 & 43.3 & 50.9 & - & - \\
				PAFPN \cite{liu2018path} & R50 & 38.1 & 58.1 & 41.3 & 19.1 & 42.5 & 54.0 & 45.2 & 209 \\
				NAS-FPN$^{\dag}$ \cite{nas-fpn} & R50 & 37.7 & 54.5 & 41.1 & 15.5 & \textbf{44.5} & \textbf{56.9} & 68.2 & 366 \\
				AugFPN \cite{augfpn} & R50 & 38.7 & 61.2 & 41.9 & 24.0 & 42.5 & 49.5 & – & – \\
				GraphFPN \cite{zhao2021graphfpn} & R50 & 39.1 & 58.3 & 39.4 & 22.4 & 38.9 & 56.7 & 100.0 & 380 \\
				FPT \cite{fpt} & R50 & 38.0 & 57.1 & 38.9 & 20.5 & 38.1 & 55.7 & 88.2 & 346 \\
				FaPN \cite{fapn} & R50 & 39.2 & - & - & 24.5 & 43.3 & 49.1 & - & - \\
				RCNet \cite{rcnet} & R50 & 40.2 & 60.9 & \textbf{43.6} & \textbf{25.0} & 43.5 & 52.9 & 68.6 & 299 \\
				AFPN \cite{yang2023afpn} & R50 & 39.0 & 57.6 & 42.1 & 19.4 & 43.0 & 55.0 & 49.8 & 194 \\
				\midrule
				\rowcolor{col}
				$A^3$-FPN-Lite (ours) & R50 & 39.7 & 60.5 & 42.8 & 23.5 & 43.2 & 56.4 & 49.6 & 210 \\
				\rowcolor{col}
				$A^3$-FPN (ours) & R50 & \textbf{40.9} & \textbf{61.7} & 43.5 & 24.6 & 44.1 & 56.7 & 76.5 & 323 \\
				\bottomrule
			\end{tabular}
		}
	\end{table}
	\subsection{Datasets and Evaluation Metrics}
	\textbf{Datasets:} We consider four widely-used benchmarks to evaluate the performance of $A^3$-FPN, including PASCAL VOC \cite{pascalvoc} and VisDrone2019-DET \cite{visdrone} for object detection, MS COCO \cite{coco} for object detection and instance segmentation, Cityscapes \cite{cityscapes} for semantic segmentation.
	
	PASCAL VOC \cite{pascalvoc} is a vital benchmark for object recognition, which has 20 object categories (e.g., vehicles, animals, household items) and contains 22136 training images (2007trainval and 2012trainval) and 4952 test images (voc2007test). The dataset is widely adopted for evaluating object detection models using mean Average Precision (mAP) at an intersection-over-union (IoU) threshold of 0.5. 
	
	VisDrone2019-DET \cite{visdrone} is a collection of aerial images captured by drones for small object detection, containing a total of 7,019 images. The dataset is split into 6,471 images for training and 548 for validation. Each image is annotated with objects from ten distinct categories: bicycle, awning tricycle, tricycle, van, bus, truck, motor, pedestrian, person, and car. The images typically have a high resolution of approximately 2000 × 1500 pixels.
	
	MS COCO \cite{coco} encompasses more than 100K images annotated for 80 object categories, which provides both per-object bounding boxes and pixel-level segmentation masks. It contains 115k images for training (train2017), 5k images for validation (val2017) and 20k images for testing (test-dev). We use the train2017 subset for training models and report model performances on the val2017 subset for comparison with SOTA methods.
	
	Cityscapes \cite{cityscapes} targets large-scale urban scene understanding for autonomous driving. It includes 5,000 high-resolution (2048×1024) images with pixel-accurate annotations, which are split into training, validation and test sets with 2975, 500 and 1525 images, respectively. The annotation consists of 30 classes (e.g., road, pedestrian, and vehicle), 19 of which are used for the semantic segmentation task.
	
	\textbf{Evaluation metrics:} For performance evaluation, Average Precision (AP) serves as the primary metric for both object detection and instance segmentation, which is calculated under the Intersection-over-Union (IoU) threshold of 0.5:0.95. We also compute $AP_s$, $AP_m$, $AP_l$ for small (area $< 32^2$ pixels), medium (\( 32^2 \leq \text{area} \leq 96^2 \) pixels), and large (area $> 96^2$ pixels) objects. Note that $AP_{box}$ and $AP_{mask}$ denote APs for bounding box and segmentation mask, respectively. For semantic segmentation, mean Intersection-over-Union (mIoU) is used as the core metric to measure the average overlap between predicted and ground-truth segmentation masks across all classes. mIoU can ensure robust performance assessment even in datasets with highly uneven class distributions.
	\begin{table}[t]
		\centering
		\caption{Small object detection performance with different feature pyramid networks on the VisDrone2019-DET dataset. The baseline model is RetinaNet \cite{focal_loss}.}
		\renewcommand{\arraystretch}{1.0}
		\label{tab:comparison_fpn_vis_drone}
		\resizebox{0.8\linewidth}{!}{
			\begin{tabular}{l|c|c|ccc|ccc}
				\arrayrulecolor{black}
				\toprule
				Method & Backbone & Epoch & AP & AP$_{50}$ & AP$_{75}$ & AP$_S$ & AP$_M$ & AP$_L$ \\
				\midrule
				RetinaNet~\cite{focal_loss} & R50 & 12 & 18.1 & 31.1 & 18.3 & 8.8 & 28.5 & 38.0 \\
				FPN~\cite{fpn} & R50 & 12 & 21.0 & 36.4 & 21.4 & 10.9 & 34.3 & 40.1 \\
				PAFPN~\cite{liu2018path} & R50 & 12 & 21.2 & 36.5 & 21.6 & 10.9 & 34.6 & 41.1 \\
				AugFPN~\cite{augfpn} & R50 & 12 & 21.7 & 37.1 & 22.2 & 11.1 & 35.4 & 40.4 \\
				FPT~\cite{fpt} & R50 & 12 & 19.3 & 33.3 & 19.2 & 9.4 & 30.0 & 38.9 \\
				RCFPN~\cite{rcnet} & R50 & 12 & 21.0 & 36.0 & 21.3 & 10.5 & 34.8 & 38.1 \\
				AFPN~\cite{yang2023afpn} & R50 & 12 & 20.7 & 36.0 & 21.2 & 10.7 & 33.4 & 36.9 \\
				CFPT \cite{cfpt} & R50 & 12 & 22.2 & 38.0 & 22.4 & 11.9 & 35.2 & 41.7\\
				\midrule
				\rowcolor{col}
				$A^3$-FPN (ours) & R50 & 12 & \textbf{23.7} & \textbf{39.4} & \textbf{24.7} & \textbf{12.4} & \textbf{37.7} & \textbf{43.8} \\
				\bottomrule
			\end{tabular}
		}
	\end{table}
	
	\subsection{Implementation Details}
	We employ MMDetection \cite{mmdet} and MMSegmentation \cite{mmseg} as the implementation platform and train models on 8 NVIDIA RTX 4090 GPUs, with 2 images processed on each. All hyperparameters in this work strictly follow the default configurations of these codebases. When integrating the proposed method into transformer-based frameworks, we replace the functionally equivalent feature fusion module with $A^3$-FPN to further prove its generalization and effectiveness. For more fair comparison, we also maintain the original training protocols from their respective official code repository.
	\subsection{Object Detection}
	\begin{figure}[t]
		\centering
		\includegraphics[width=0.9\linewidth]{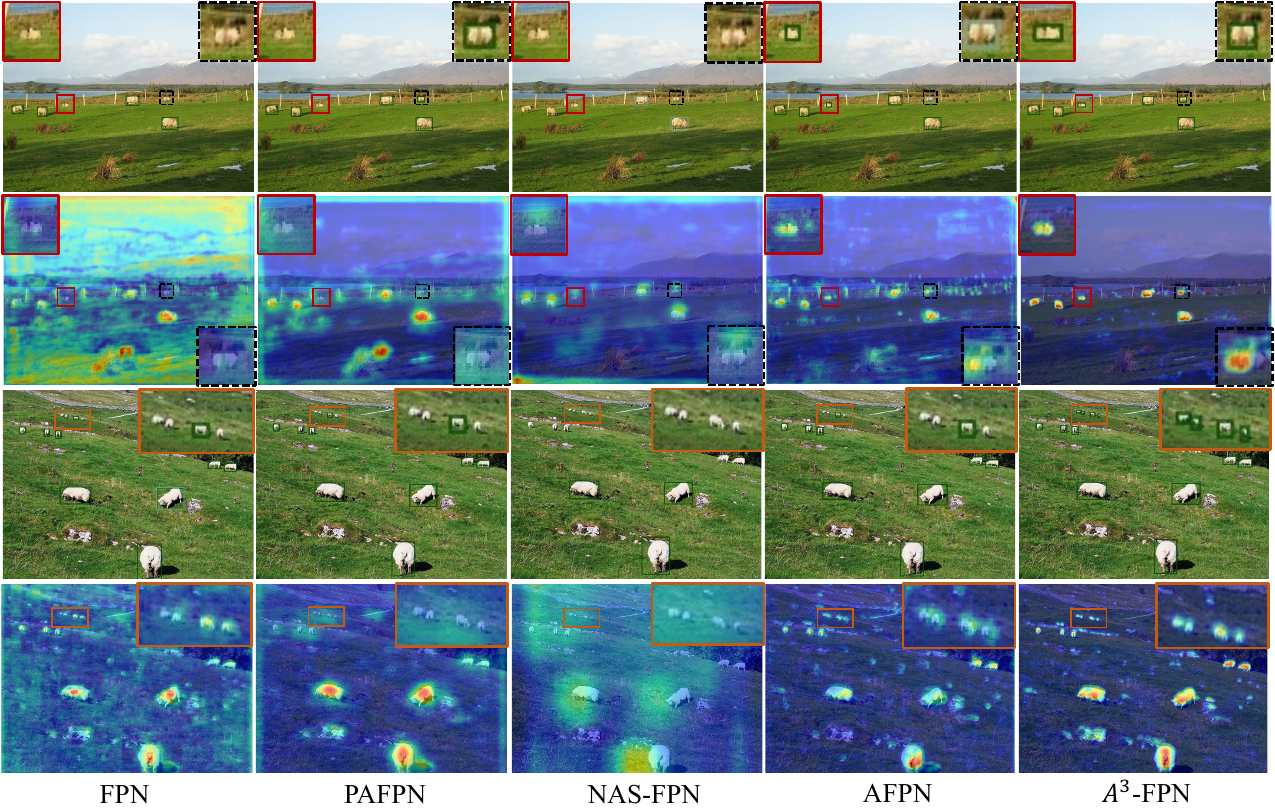}
		\caption{Qualitative evaluation of various feature pyramid networks for object detection on MS COCO validation set, including FPN \cite{fpn}, PAFPN \cite{liu2018path}, NAS-FPN \cite{nas-fpn}, AFPN \cite{yang2023afpn} and our $A^3$-FPN. Odd rows are the detection results, and the others are the corresponding AblationCAM \cite{ablation_cam} visualization. The object category in the images is sheep.}
		\label{fig:detect}
	\end{figure}
	\noindent
	\begin{table}[t]
		\centering
		\caption{Performance comparison between state-of-the-art detectors with and without incorporating $A^3$-FPN on MS COCO dataset. CNN detectors are evaluated on COCO {\tt test-dev}, while vision transformer detectors are tested on COCO {\tt val2017}.}
		\renewcommand{\arraystretch}{1.2}
		\label{tab:sota_det}
		\resizebox{1.0\linewidth}{!}{
			\begin{tabular}{l|l|l|c|ccc|ccc}
				\toprule
				Type & Method  & Backbone & Epoch & AP & AP$_{50}$ & AP$_{75}$ & AP$_{\it S}$ & AP$_{\it M}$ & AP$_{\it L}$ \\
				\midrule
				\multirow{7}{*}{\textit{\makecell[c]{multi-stage \\ CNN detectors}}} 
				& Cascade R-CNN \cite{cai2018cascade} & R101  & 18 & 42.8 & 62.1 & 46.3 & 23.7 & 45.5 & 55.2 \\
				& $A^3$-FPN-Lite & R101 & 18 & 44.3 & 62.3 & 47.6 & 25.3 & 47.1 & 57.6 \\
				& $A^3$-FPN & R101 & 18 & 45.1 & 62.9 & 48.5 & 26.2 & 47.9 & 58.1 \\
				\cdashline{2-10}
				& RepPointsV2 \cite{reppointsv2} & R101 & 24 & 46.0 & 65.3 & 49.5 & 27.4 & 48.9 & 57.3 \\
				& $A^3$-FPN-Lite & R101 & 24 & 47.3 & 66.1 & 50.4 & 27.9 & 49.6 & 58.4  \\
				& $A^3$-FPN & R101 & 24 & 48.2 & 67.2 & 51.1 & 28.5 & 50.9 & 60.3  \\
				\midrule
				\multirow{6}{*}{\textit{\makecell[c]{one-stage \\ CNN detectors}}}
				& FCOS \cite{fcos} & R101 & 24 & 41.5 & 60.7 & 45.0 & 24.4 & 44.8 & 51.6 \\
				& $A^3$-FPN-Lite & R101 & 24 & 43.4 & 62.8 & 47.1 & 26.2 & 45.9 & 52.9 \\
				& $A^3$-FPN & R101 & 24 & 44.7 & 63.9 & 48.2 & 27.8 & 47.1 & 54.3 \\
				\cdashline{2-10}
				& GFLV2 \cite{gflv2} & R101 & 24 & 46.2 & 64.3 & 50.5 & 27.8 & 49.9 & 57.0 \\
				& $A^3$-FPN-Lite & R101 & 24 & 47.1 & 65.3 & 51.2 & 28.2 & 51.3 & 58.6 \\
				& $A^3$-FPN & R101 & 24 & 48.2 & 66.2 & 52.8 & 28.8 & 51.9 & 60.2 \\
				\midrule
				\multirow{8}{*}{\textit{\makecell[c]{vision Transformer \\ detectors \\ (300 queries)}}}
				& RT-DETR \cite{rtdetrv1} & R50 & 72 & 53.1 & 71.3 & 57.7 & 34.8 & 58.0 & 70.0 \\
				& +$A^3$-FPN-Lite & R50 & 72 & 54.1 & 71.9 & 58.3 & 35.4 & 58.3 & 71.2 \\
				& +$A^3$-FPN & R50 & 72 & 54.9 & 72.6 & 58.8 & 36.1 & 58.7 & 72.3 \\
				\cdashline{2-10}
				& D-FINE \cite{dfine} & HGNetv2-L & 80 & 54.0 & 71.6 & 58.4 & 36.5 & 58.0 & 71.9 \\
				& +$A^3$-FPN-Lite & HGNetv2-L & 80 & 54.8 & 72.3 & 59.1 & 36.8 & 58.9 & 72.5 \\
				& +$A^3$-FPN & HGNetv2-L & 80 & 55.5 & 72.9 & 60.0 & 37.2 & 59.4 & 73.1 \\
				\cdashline{2-10}
				& Mask DINO \cite{mask_dino} & R50 & 50 & 50.5 & - & - & - & - & - \\
				& +$A^3$-FPN & R50 & 50 & 51.6 & - & - & - & - & - \\
				\bottomrule
			\end{tabular}
		}
	\end{table}
	We first compare the performance of Faster RCNN with different multi-scale feature fusion methods under the same training schedule. As shown in Table \ref{tab:comparison_fpn}, $A^3$-FPN achieves a leading position in AP, AP$_{50}$, and AP$_S$ with 76.5 M parameters and 323 GFLOPs, outperforming the other pyramid networks in the table. Our lightweight $A^3$-FPN-Lite maintains relatively lower computational cost and inference latency (Fig. \ref{fig:flops_accuracy}) and attains 39.7 AP and 60.5 AP$_{50}$. Additionally, we also present the qualitative evaluation between our method and other feature pyramid networks in Fig. \ref{fig:detect}, including the detection visualization and corresponding AblationCAM \cite{ablation_cam}. Fig. \ref{fig:detect} implies that coarse sampling and vanilla fusion operations will produce inconsistent and less expressive features, finally resulting in misclassification and underdetection. With MCAtten and ICAtten, $A^3$-FPN learns more discriminative and intra-category similar features and reduces intra-category inconsistency in the fusion stage. Thus, our model is more focused on object pixel regions, and correctly predicts the locations and class of all \textit{sheep} objects. Moreover, we also evaluate our model on VisDrone2019-DET to analyze its performance in smaller goal-intensive scenarios. In Table \ref{tab:comparison_fpn_vis_drone}, our method improves the baseline by +5.6 AP and outperforms the other competing methods across all reported metrics. These results further confirm that $A^3$-FPN effectively filters redundant information and enhances discriminative feature learning for small and densely clustered instances in complex environments.
	
	To further demonstrate the versatility of our approach, we incorporate $A^3$-FPN into leading CNN (multi-stage and one-stage) and Transformer detectors. As summarized in Table \ref{tab:sota_det}, extensive experiments indicate that our method can remarkably boost the performance of both detector frameworks. By incorporating $A^3$-FPN-Lite and $A^3$-FPN, Cascade R-CNN \cite{cai2018cascade} achieves 1.5 AP and 2.3 AP improvements, respectively. When adopting $A^3$-FPN-Lite and $A^3$-FPN to learn multi-scale features, the point representation-based detector RepPointsV2 \cite{reppointsv2} surpasses the original baseline performance by 1.3 AP and 2.2 AP. In the one-stage detector, $A^3$-FPN increases the detection precision of FCOS \cite{fcos} and GFLV2 \cite{gflv2} by 3.2 AP and 2.0 AP, while $A^3$-FPN-Lite enhances them by 1.9 AP and 0.9 AP. For the detection transformer, replacing Hybrid Encoder with $A^3$-FPN-Lite and $A^3$-FPN leads to AP improvements of \textbf{+1.0} and \textbf{+1.8} for RT-DETR \cite{rtdetrv1}, and \textbf{+0.8} and \textbf{+1.5} for D-FINE \cite{dfine}, respectively. In addition, integrating $A^3$-FPN into the unified visual multi-task framework further validates its effectiveness, with Mask DINO \cite{mask_dino} achieving a \textbf{+1.1 AP} enhancement. These improvements also indicate that intra-category inconsistency and object displacement in the feature fusion are still widely prevalent challenges in advanced models.
	\subsection{Instance Segmentation}
\begin{figure}[t]
		\centering
		\includegraphics[width=1.0\linewidth]{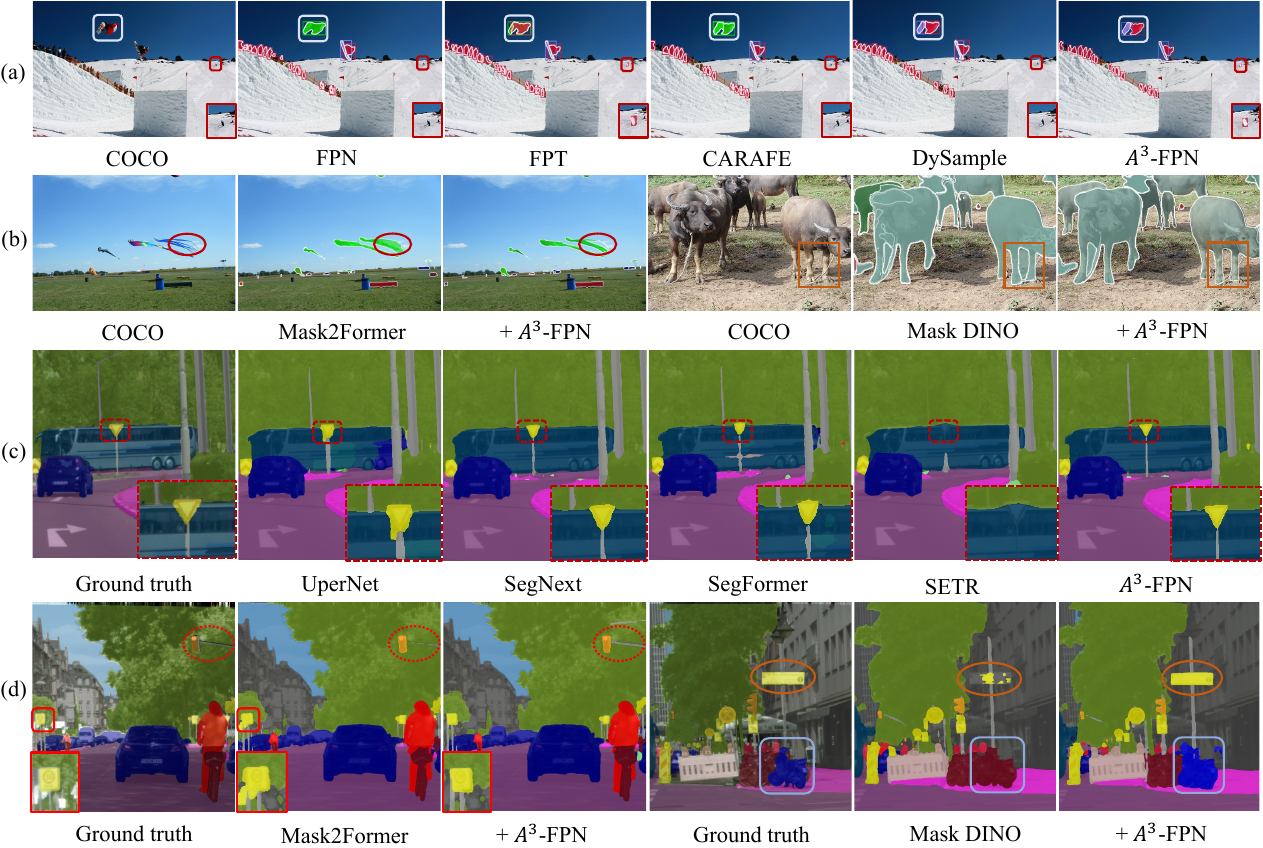}
		\caption{Qualitative evaluation. (a) Instance segmentation results of Mask RCNN \cite{he2017mask} with different feature fusion approaches, including FPN \cite{fpn}, FPT \cite{fpt}, DySample \cite{dysample}, CARAFE \cite{carafe++} and our $A^3$-FPN; (b) and (d) are comparison between some unified transformer-based models (Mask2Former \cite{mask2former} and Mask DINO \cite{mask_dino}) with and without integrating $A^3$-FPN on the instance and semantic segmentation task respectively; (c) Semantic segmentation visualization on Cityscapes validation set using different semantic segmentors, which are UperNet \cite{upernet}, SegNext \cite{segnext}, SegFormer \cite{segformer}, SETR \cite{setr}, and our $A^3$-FPN. Additional qualitative results are presented in Appendix E and F.}
		\label{fig:seg}
	\end{figure}

	\begin{table}[t]
		\centering
		\caption{Mask AP comparison of different instance segmentation models on COCO {\tt val2017}. CNN methods adopt Mask R-CNN \cite{he2017mask} as the baseline architecture. $^{\textbf{*}}$ denotes mask AP is evaluated on instance ground truths derived from panoptic annotations \cite{oneformer} and $^{\dag}$ indicates re-implementation results.}
		\renewcommand{\arraystretch}{1.2}
		\label{tab:sota_instance_seg}
		\resizebox{1.0\linewidth}{!}{
			\begin{tabular}{l|l|l|c|ccc|ccc}
				\toprule
				Type & Method  & Backbone & Epoch & AP & AP$_{50}$ & AP$_{75}$ & AP$_{\it S}$ & AP$_{\it M}$ & AP$_{\it L}$ \\
				\midrule
				\multirow{10}{*}{\makecell[c]{\textit{CNN methods}}}
				& FPN \cite{fpn} & R50  & 12 & 34.7 & 55.7 & 37.2 & 18.3 & 37.4 & 47.2 \\
				& FPT \cite{fpt} & R50  & 12 & 36.8 & 55.9 & \textbf{38.6} & 18.8 & 35.3 & \textbf{54.2} \\
				& CARAFE \cite{carafe++} & R50  & 12 & 35.4 & 56.7 & 37.6 & 16.9 & 38.1 & 51.3 \\
				& DySample+ \cite{dysample} & R50  & 12 & 35.7 & 57.3 & 38.2 & 17.3 & 38.2 & 51.8 \\
				& FreqFusion \cite{frequency_aware} & R50  & 12 & 36.0 & 57.9 & 38.1 & 17.9 & 39.0 & 52.3 \\
				& $A^2$-FPN \cite{a2-fpn} & R50  & 12 & 36.2 & 58.4 & 38.1 & 20.1 & \textbf{39.2} & 49.6 \\				
				& TFPN$^{\dag}$ \cite{TFPN} & R50  & 12 & 36.6 & 58.5 & 38.0 & 19.7 & 39.0 & 52.5 \\
				& $A^3$-FPN-Lite & R50 & 12 & 36.3 & 57.6 & 38.2 & 19.5 & 38.6 & 52.9 \\
				& $A^3$-FPN & R50 & 12 & \textbf{37.1} & \textbf{58.7} & 38.4 & \textbf{20.2} & 39.1 & 53.7 \\
				\cdashline{2-10}
				& DynaMask \cite{dynamask} & R50 & 12 & 37.6 & 57.4 & 40.5 & 20.7 & 40.4 & 50.3 \\
				& $A^3$-FPN-Lite & R50 & 12 & 38.1 & 57.8 & 41.2 & 21.1 & 40.9 & 51.6  \\
				& $A^3$-FPN & R50 & 12 & 38.8 & 58.4 & 42.0 & 21.7 & 41.3 & 52.5  \\
				\midrule
				\multirow{6}{*}{\makecell[c]{\textit{Transformer methods}}}
				& Mask2Former \cite{mask2former} & R50 & 50 & 43.7 & 66.0 & 46.9 & 23.4 & 47.2 & 64.8 \\
				& +$A^3$-FPN & R50 & 50 & 44.2 & 66.5 & 47.9 & 24.5 & 47.6 & 65.6 \\
				\cdashline{2-10}
				& OneFormer \cite{oneformer} & Swin-L & 100 & 49.0$^{\textbf{*}}$ & - & - & - & - & - \\
				& +$A^3$-FPN & Swin-L & 100 & 49.6$^{\textbf{*}}$ & - & - & - & - & - \\
				\cdashline{2-10}
				& Mask DINO \cite{mask_dino} & R50 & 50 & 45.4 & 67.9 & 49.3 & 25.2 & 48.3 & 65.8 \\
				& +$A^3$-FPN & R50 & 50 & 46.2 & 68.7 & 50.2 & 26.1 & 48.9 & 66.6 \\
				\bottomrule
			\end{tabular}
		}
	\end{table}
	Quantitative evaluations of instance segmentation are summarized in Table \ref{tab:sota_instance_seg}. With ResNet-50 \cite{resnet} and Mask R-CNN \cite{he2017mask} serving as the backbone and segmentation head, our $A^3$-FPN showcases noteworthy performance on COCO {\tt val2017}, surpassing the other methods in the table across AP, AP$_{50}$, and AP$_{\it S}$. Notably, $A^3$-FPN-Lite attains 36.3 mask AP with comparable computational efficiency (similar parameters and FLOPs to lightweight counterparts), outperforming CARAFE \cite{carafe++} by 0.9 AP, DySample \cite{dysample} by 0.6 AP, and FreqFusion \cite{frequency_aware} by 0.3 AP. Qualitative results in Fig. \ref{fig:seg} (a) and (b) demonstrate $A^3$-FPN's outstanding segmentation and classification precision compared to conventional FPNs and upsampling methods, which can be attributed to the content-aware attention mechanism that effectively decreases false positives and missing detections. Furthermore, integrating $A^3$-FPN-Lite and $A^3$-FPN into DynaMask \cite{dynamask} elevates baseline performance by 0.5 AP and 1.2 AP, respectively. When applied to transformer-based instance segmentors, our method delivers consistent gains: Mask2Former (+0.5 AP), OneFormer (+0.6 AP), and Mask DINO (+0.8 AP), demonstrating its versatility in multi-scale feature representation.
	
	\subsection{Semantic Segmentation}
	\begin{table}[t]
		\centering
		\caption{Comparison with recent state-of-the-art semantic segmentation methods on Cityscapes \cite{cityscapes} validation set. We calculate mIoU to assess the model performance and intra-category consistency of the final predictions. $^{\dag\dag}$ indicates our re-implementation results, while s.s. and m.s. denotes single-scale and multi-scale training, respectively.}
		\renewcommand{\arraystretch}{1.2}
		\label{tab:sota_sem_seg}
		\resizebox{1.0\linewidth}{!}{
			\begin{tabular}{l|l|l|l|c|c|c}
				\toprule
				Type & Method  & Backbone & Crop Size & Schedule & mIoU (s.s.) & mIoU (m.s.) \\
				\midrule
				\multirow{10}{*}{\textit{\makecell[c]{CNN \\ segmentors}}}
				& PointRend \cite{pointrend} & R50 & 512$\times$1024 & 80k & 76.47 & 78.13 \\
				& FaPN$^{\dag\dag}$ \cite{fapn} & R50 & 512$\times$1024 & 80k & 79.07 & - \\
				& $A^3$-FPN-Lite & R50 & 512$\times$1024 & 80k & 78.98 & 80.48 \\
				& $A^3$-FPN & R50 & 512$\times$1024 & 80k & 79.93 & 81.54 \\
				\cdashline{2-7}
				& PSPNet \cite{pspnet} & R50 & 512$\times$1024 & 80k & 78.55 & 79.79 \\
				& PSANet \cite{psanet} & R50 & 512$\times$1024 & 80k & 77.24 & 78.69 \\ 
				& SegNeXt \cite{segnext} & MSCAN-T & 1024$\times$1024 & 160k & 79.80 & 81.40 \\                
				\cdashline{2-7}
				& UperNet \cite{upernet} & R50 & 512$\times$1024 & 80k & 78.19 & 79.19 \\
				& $A^3$-FPN-Lite & R50 & 512$\times$1024 & 80k & 78.93 & 80.36 \\
				& $A^3$-FPN & R50 & 512$\times$1024 & 80k & 79.65 & 81.22 \\
				\midrule
				\multirow{14}{*}{\textit{\makecell[c]{Transformer \\ segmentors}}}
				& SETR PUP \cite{setr} & ViT-L & 768$\times$768 & 80k & 79.2 & 81.0 \\
				& Segmenter \cite{segmenter} & ViT-L-16 & 768$\times$768 & 80k & 79.1 & 81.3 \\
				\cdashline{2-7}
				& SegFormer \cite{segformer} & MIT-B1 & 1024$\times$1024 & 160k & 78.6 & 79.7 \\
				& +FreqFusion \cite{frequency_aware} & MIT-B1 & 1024$\times$1024 & 160k & 80.1 & - \\
				& +$A^3$-FPN & MIT-B1 & 1024$\times$1024 & 160k & 80.8 & - \\
				\cdashline{2-7}
				& Mask2Former \cite{mask2former} & R50 & 512$\times$1024 & 90k & 79.4 & 82.2 \\
				& +FreqFusion \cite{frequency_aware} & R50 & 512$\times$1024 & 90k & 80.5 & - \\
				& +$A^3$-FPN & R50 & 512$\times$1024 & 90k & 81.1 & - \\
				\cdashline{2-7}
				& Mask DINO \cite{mask_dino} & R50 & 512$\times$1024 & 90k & 79.8 & - \\
				& +$A^3$-FPN & R50 & 512$\times$1024 & 90k & 81.6 & - \\
				\cdashline{2-7}
				& OneFormer \cite{oneformer} & Swin-L & 512$\times$1024 & 90k & 83.0 & 84.4 \\
				& +$A^3$-FPN & Swin-L & 512$\times$1024 & 90k & 83.9 & 85.6 \\
				\bottomrule
			\end{tabular}
		}
	\end{table}
	
	As shown in Table \ref{tab:sota_sem_seg} and Fig. \ref{fig:seg} (c) and (d), $A^3$-FPN exhibits prominent advantages in both CNN and transformer-based semantic segmentors on the Cityscapes validation set. When using ResNet-50 as the backbone and PointRend \cite{pointrend} as the mask head, $A^3$-FPN achieves 79.93 and 81.54 mIoU under single-scale and multi-scale training, while $A^3$-FPN-Lite attains 78.98/80.48 (s.s./m.s.) mIoU with lower parameter counts. Additionally, compared to classical FPN-based methods like UperNet \cite{upernet}, $A^3$-FPN and $A^3$-FPN-Lite achieve a significant gain of 1.46 and 0.74 mIoU, respectively.
	When integrated with transformer-based segmentors, $A^3$-FPN delivers consistent performance gains across multiple architectures: SegFormer \cite{segformer} and Mask2Former \cite{mask2former} reach \textbf{80.8} and \textbf{81.1} mIoU respectively, surpassing their baseline implementations by \textbf{+2.2} and \textbf{+1.7}, and the FreqFusion-enhanced counterparts \cite{frequency_aware} by \textbf{+0.7} and \textbf{+0.6}; Mask DINO \cite{mask_dino} and OneFormer \cite{oneformer} exhibit improvements of 1.8 and 0.9 mIoU respectively, demonstrating broad compatibility of $A^3$-FPN. All these results strongly underscore the robustness and efficacy of $A^3$-FPN in advancing semantic segmentation performance. 
	
	\begin{table}[t]
		\centering
		\caption{Ablation study for different feature fusion frameworks. AP$^{\text{box}}$ and AP$^{\text{mask}}$ are evaluated on MS COCO {\tt val2017}, while mIoU is calculated on Cityscapes.}
		\renewcommand{\arraystretch}{1.0}
		\label{tab:ablation_framework}
		\resizebox{0.75\linewidth}{!}{
			\begin{tabular}{l|ccc}
				\toprule
				Framework  & AP$^{\text{box}}$ & AP$^{\text{mask}}$ & mIoU \\
				\midrule
				\textit{with FPN's modules:}  & & & \\
				top-down framework \cite{fpn} & 38.2  & 34.7 & -  \\
				\makecell[l]{top-down + bottom up framework \cite{liu2018path}}  & 38.7  & 35.1 & - \\
				Gather-and-Distribute framework \cite{goldyolo} & 39.2  & 35.4 & - \\
				\hdashline
				\makecell[l]{top-down asymptotically disentangled framework}  & 39.8 & 35.9 & - \\ 
				\midrule
				
				\textit{with $A^3$-FPN's modules:}   & & &  \\
				top-down framework & 39.3  & 35.8 & -  \\
				\makecell[l]{top-down + bottom up framework}  & 39.9  & 36.1 & -  \\
				Gather-and-Distribute framework & 40.4  & 36.5 & -  \\
				\hdashline
				\makecell[l]{top-down asymptotically disentangled framework}  & 41.9  & 37.3 & 79.11  \\ 
				\makecell[l]{bottom-up asymptotically disentangled framework}  & 41.3  & 36.9 & 79.65 \\ 
				\bottomrule
			\end{tabular}
		}
	\end{table}
	
	\begin{table}[t]
		\centering
		\begin{minipage}{0.46\textwidth}
			\centering
			\captionof{table}{Performance comparison with other sampling methods and results are reported on COCO {\tt val2017}.}
			\label{tab:sampling_ablation}
			\resizebox{0.8\linewidth}{!}{
				\begin{tabular}{l |cc}
					\toprule
					\textbf{Sampler} & AP$^{\text{box}}$ & AP$^{\text{mask}}$ \\
					\midrule
					Nearest  & 38.3 & 34.7 \\
					PixelShuffle \cite{pixel_shuffle}  & 38.5  & 34.8 \\
					CARAFE \cite{carafe++}  & 39.2  & 35.4 \\
					A2U \cite{a2u} & 38.2  & 34.6 \\
					SAPA-B \cite{sapa} & 38.7  & 35.1 \\
					FADE \cite{fade} & 39.1  & 35.1 \\
					DySample+ \cite{dysample}  & 39.6  & 35.7 \\
					\hdashline
					\makecell[l]{Resampler (ours)} & 40.2  & 36.1 \\
					\bottomrule
				\end{tabular}
			}
		\end{minipage}
		\hfill 
		\begin{minipage}{0.45\textwidth}
			\centering
			\captionof{table}{Performance comparison with other feature fusion methods and results are reported on COCO {\tt val2017}.}
			\label{tab:fusion_ablation}
			\resizebox{1.0\linewidth}{!}{
				\begin{tabular}{l |ccc}
					\toprule
					\textbf{Method} & AP$^{\text{box}}$ & AP$_{50}^{\text{box}}$ \\
					\midrule
					Concatenation+Conv & 37.7  & 57.4 \\
					Sum+Conv & 37.4  & 57.3 \\
					Adaptive Spatial Fusion \cite{yang2023afpn} & 37.9  & 57.7 \\
					Channel Attention \cite{a2-fpn} & 38.2  & 58.1 \\
					\hdashline
					\makecell[l]{Context Weight Generator (ours)} & 38.6  & 58.6 \\
					\bottomrule
				\end{tabular}
			}
		\end{minipage}
		
		\vspace{0.5em}
		\small
	\end{table}
	
	\subsection{Ablation Studies}
	\label{ablation}
	In this section, we perform thorough ablation studies for the proposed method, including threshold selection in ICAtten, group division in the grouping strategy, validating the effectiveness of the asymptotically disentangled framework, and systematic evaluation of feature sampling and fusion operations in $A^3$-FPN.
	
	\textbf{Feature fusion framework.} Feature fusion frameworks play a critical role in multi-scale feature learning. To validate the framework superiority of $A^3$-FPN, we conduct comparative experiments on Mask R-CNN using different multi-scale fusion frameworks, including layer-wise framework (top-down \cite{fpn} and top-down + bottom-up \cite{liu2018path}), global fusion framework (Gather-and-Distribute in Gold-YOLO \cite{goldyolo}) and the asymptotically disentangled framework. To ensure a fair comparison and eliminate biases from well-designed fusion modules, we implement the standard FPN's and $A^3$-FPN's modules across all frameworks. Quantitative results in Table~\ref{tab:ablation_framework} show that, regardless of the sampling and fusion methods employed, the proposed asymptotically disentangled framework consistently outperforms other fusion frameworks across object detection, instance segmentation, and semantic segmentation tasks. This superiority stems from its multi-column nesting paradigm and column-wise disentangled mechanism, which collectively enhance the model's capacity to learn both discriminative feature representations and semantically rich embeddings.
	
	Table~\ref{tab:ablation_framework} further reveals the task-specific advantages of the proposed asymptotic framework. Top-down asymptotic design achieves superior performance in position-sensitive tasks (e.g., object detection, instance segmentation), as its hierarchical feature propagation enables deeper semantic enrichment through more column transformations. This design prioritizes the flow of high-level semantic content (e.g., object existence and category) to lower levels via column-wise spread, enhancing localization accuracy while maintaining class consistency. Bottom-up asymptotic design excels in semantic segmentation, where accurate boundary delineation critically depends on semantically rich low-level features. By progressively enriching low-level features through 3-column nesting, the bottom-up paradigm facilitates the integration of semantic information at the low level and fine-grained pixel-level mask learning, improving semantic segmentation precision.
	
	\begin{table}[t]
		\caption{Ablation study for MCAtten and ICAtten. With Faster RCNN \cite{faster_rcnn} as the detector, models are trained and evaluated on PASCAL VOC training set (2007trainval and 2012trainval) and voc2007test, respectively.}
		\label{tab:ablation_module}
		\centering
		\renewcommand{\arraystretch}{1.0}
		\resizebox{0.75\linewidth}{!}{
			\begin{tabular}{c c c| c c c}
				\toprule
				\multicolumn{2}{c}{\textbf{MCAtten}}
				& \multirow{2}{*}{\textbf{ICAtten}}
				& \multirow{2}{*}{AP}
				& \multirow{2}{*}{AP$_{50}$}
				& \multirow{2}{*}{AP$_{75}$}
				\\
				\cmidrule(lr){1-2}
				\makecell[c]{context-aware resampler} & \makecell[c]{context weight generator} 
				&  &  &  & 
				\\
				\hline
				\multicolumn{3}{c|}{\textbf{standard feature fusion}} & 54.13 & 80.23 & 59.56 \\
				\hdashline
				\colorbox[HTML]{ECF4FF}{\ding{51}} & \ding{55} & \ding{55} & 55.47 & 82.43 & 60.97 \\
				\colorbox[HTML]{ECF4FF}{\ding{51}} & \colorbox[HTML]{ECF4FF}{\ding{51}} & \ding{55} & 56.45 & 82.27 & 62.03 \\
				\colorbox[HTML]{ECF4FF}{\ding{51}} & \colorbox[HTML]{ECF4FF}{\ding{51}} & \colorbox[HTML]{ECF4FF}{\ding{51}} & 57.08 & 82.56 & 62.13 \\
				\bottomrule
			\end{tabular}
		}
	\end{table}
	
	\textbf{Feature sampling and fusion operations.}
	We first validate the efficacy of the proposed MCAtten and ICAtten in $A^3$-FPN. As demonstrated in Table~\ref{tab:ablation_framework}, frameworks incorporating MCAtten and ICAtten consistently outperform those using standard FPN, achieving superior AP$^{\text{box}}$, AP$^{\text{mask}}$ and mIoU. To further dissect the superiority of individual components, we conduct a detailed analysis of the context-aware resampler and context weight generator. Compared to recent sampling methods, our context-aware resampler exhibits distinct advantages. As shown in Table~\ref{tab:sampling_ablation}, when integrated with Mask R-CNN and ResNet-50 for instance segmentation, the proposed context-aware resampler outperforms the second-place Dysample \cite{dysample} by remarkable margins of 0.6 AP$^{\text{box}}$ and 0.4 AP$^{\text{mask}}$. Different from context-agnostic and independent sampling methods like DySample \cite{dysample} and CARAFE \cite{carafe++}, our resampler depends on multi-scale feature hierarchies and uses adjacent-level information to enhance sampling quality. Similarly, when evaluated with Faster R-CNN and ResNet-50 (Table~\ref{tab:fusion_ablation}), our feature interaction strategy by the context weight generator surpasses the other feature fusion approaches, achieving 0.7 AP$^{\text{box}}$ and 0.4 AP$^{\text{box}}$ improvements over Adaptive Spatial Fusion \cite{yang2023afpn} and Channel Attention \cite{a2-fpn}, respectively. Finally, we investigate the cooperation effects of combining all proposed modules. As summarized in Table~\ref{tab:ablation_module} and Figure \ref{fig:ablation_module}, only the integration of all components yields optimal performance across AP, AP$_{50}$ and AP$_{75}$, which further confirms the contribution and necessity of each component in $A^3$-FPN.
	\begin{table}[t]
		\centering
		\caption{APs on PASCAL VOC 2007 test set and mIoUs on Cityscapes validation set with different thresholds in ICAtten.}
		\label{tab:ablation_threshold}
		\renewcommand{\arraystretch}{1.0}
		\small
		\resizebox{0.6\linewidth}{!}{
			\begin{tabular}{l|ccccccccc}
				\toprule
				\diagbox{\textbf{Metric}}{\textbf{Threshold}} & 0.2 & 0.3 & 0.4 & 0.5 & 0.6 & 0.7 & 0.8 \\
				\midrule
				AP & 56.53 & 56.65 & 56.36 & 57.08 & 56.73 & 56.51 & 56.45 \\
				mIoU & 79.15 & 79.29 & 79.57 & 79.65 & 79.45 & 79.34 & 79.23 \\
				\bottomrule
			\end{tabular}
		}
	\end{table}
	\begin{table}[t]
		\centering
		\caption{APs of our method on PASCAL VOC 2007 test set with different channel numbers per group in the offset generator and ICAtten.}
		\label{tab:ablation_group}
		\renewcommand{\arraystretch}{1.0}
		\small
		\resizebox{0.6\linewidth}{!}{
			\begin{tabular}{l|ccccccc}
				\toprule
				\diagbox{\textbf{Module}}{\textbf{Channels}} & 4 & 8 & 16 & 32 & 64 \\
				\midrule
				offset generator & 56.47  & 56.86 & 57.08 & 56.98 & 56.79 \\
				ICAtten & 56.34  & 56.95 & 57.08 & 56.69 & 56.91 \\
				\bottomrule
			\end{tabular}
		}
	\end{table}
	
	\textbf{Threshold setting in ICAtten.} In ICAtten, the higher threshold means the stricter information compression and filtering. To explore the effect of different threshold settings, we vary the threshold from 0.2 to 0.8 gradually to compare APs on PASCAL VOC and mIoUs on Cityscapes. As illustrated in Table \ref{tab:ablation_threshold}, the optimal AP-mIoU emerges at a threshold of 0.5, where AP reaches 57.08 and mIoU remains competitive at 79.65. 
	Therefore, we adopt the optimal threshold of 0.5 for ICAtten.
	
	\textbf{Group division in the offset generator and ICAtten.} To determine the optimal number of channels in dividing into groups, we conduct ablation experiments with varying numbers of channels and record the APs on PASCAL VOC, as shown in Table \ref{tab:ablation_group}. The results show that 16 channels per group achieves the optimal AP for both offset generator and ICAtten, which is consistent with the observations reported in \cite{gn}. Accordingly, we use this setting in all subsequent experiments.

	\section{Conclusion}
	In this work, we rethink the existing multi-scale feature fusion methods for dense visual prediction tasks, analyzing their defects concerning the framework design and fusion operations. Building on this, we propose $A^3$-FPN, an asymptotic content-aware pyramid attention network, which integrates the asymptotically disentangled framework with content-aware attention-based fusion operations. $A^3$-FPN efficiently resolves intra-category inconsistencies and boundary feature displacement for medium-to-large objects while enhancing discriminative feature learning for small and densely clustered instances. Despite its outstanding performance, we have not yet systematically examined $A^3$-FPN under more extreme or challenging scenarios, such as out-of-distribution inputs, low-light conditions, or severe occlusion. These situations are common in real-world applications and may significantly affect the robustness of the proposed approach. We leave this as an important direction for future work, where we plan to explore model adaptation and enhancement strategies to improve performance in these challenging environments.
	
	\bibliographystyle{elsarticle-num} 
	\biboptions{sort&compress}
	\bibliography{ref}

@INPROCEEDINGS{visdrone,
  author={Du, Dawei and Zhu, Pengfei and Wen, Longyin and Bian, Xiao and et al.},
  booktitle={2019 IEEE/CVF International Conference on Computer Vision Workshop (ICCVW)}, 
  title={VisDrone-DET2019: The Vision Meets Drone Object Detection in Image Challenge Results}, 
  year={2019},
  volume={},
  number={},
  pages={213-226},
}

@ARTICLE{cfpt,
  author={Du, Zewen and Hu, Zhenjiang and Zhao, Guiyu and Jin, Ying and Ma, Hongbin},
  journal={IEEE Transactions on Geoscience and Remote Sensing}, 
  title={Cross-Layer Feature Pyramid Transformer for Small Object Detection in Aerial Images}, 
  year={2025},
  volume={63},
  number={},
  pages={1-14},
}

@ARTICLE{TFPN,
  author={Liu, Dongfang and Liang, James and Geng, Tony and Loui, Alexander and Zhou, Tianfei},
  journal={IEEE Transactions on Image Processing}, 
  title={Tripartite Feature Enhanced Pyramid Network for Dense Prediction}, 
  year={2023},
  volume={32},
  number={},
  pages={2678-2692},
}

@INPROCEEDINGS{lr-fpn,
  author={Li, Hanqian and Zhang, Ruinan and Pan, Ye and Ren, Junchi and Shen, Fei},
  booktitle={2024 International Joint Conference on Neural Networks (IJCNN)}, 
  title={LR-FPN: Enhancing Remote Sensing Object Detection with Location Refined Feature Pyramid Network}, 
  year={2024},
  volume={},
  number={},
  pages={1-8},
}

@article{enhancing_sfi,
  title={Enhancing aerial object detection with selective frequency interaction network},
  author={Weng, Weijie and Wei, Mengwan and Ren, Junchi and Shen, Fei},
  journal={IEEE Transactions on Artificial Intelligence},
  volume={5},
  number={12},
  pages={6109--6120},
  year={2024},
  publisher={IEEE}
}

@article{sigmoid_approximation,
  title={Approximation by superpositions of a sigmoidal function},
  author={Cybenko, George},
  journal={Mathematics of control, signals and systems},
  volume={2},
  number={4},
  pages={303--314},
  year={1989},
  publisher={Springer}
}

@inproceedings{mlp_width,
  title={The expressive power of neural networks: A view from the width},
  author={Lu, Zhou and Pu, Hongming and Wang, Feicheng and Hu, Zhiqiang and Wang, Liwei},
  booktitle={Advances in neural information processing systems},
  pages={6232–6240},
  year={2017}
}

@article{cednet_pr,
  title={CEDNet: A cascade encoder--decoder network for dense prediction},
  author={Zhang, Gang and Li, Ziyi and Tang, Chufeng and Li, Jianmin and Hu, Xiaolin},
  journal={Pattern Recognition},
  volume={158},
  pages={111072},
  year={2025},
  publisher={Elsevier}
}

@article{dense_pr,
  title={Rethinking local and global feature representation for dense prediction},
  author={Chen, Mohan and Zhang, Li and Feng, Rui and Xue, Xiangyang and Feng, Jianfeng},
  journal={Pattern Recognition},
  volume={135},
  pages={109168},
  year={2023},
  publisher={Elsevier}
}

@article{small_pr,
title = {Real-time small object detection using adaptive weighted fusion of efficient positional features},
journal = {Pattern Recognition},
volume = {167},
pages = {111717},
year = {2025},
issn = {0031-3203},
author = {Xin Ding and Ruichen Zhang and Qiong Liu and You Yang},
}

@article{yang2023afpn,
  title={Asymptotic feature pyramid network for labeling pixels and regions},
  author={Yang, Guoyu and Lei, Jie and Tian, Hao and Feng, Zunlei and Liang, Ronghua},
  journal={IEEE Transactions on Circuits and Systems for Video Technology},
  year={2024},
  volume={34},
  number={9},
  pages={7820-7829},
  publisher={IEEE}
}

@inproceedings{pixel_shuffle,
  title={Real-time single image and video super-resolution using an efficient sub-pixel convolutional neural network},
  author={Shi, Wenzhe and Caballero, Jose and Husz{\'a}r, Ferenc and Totz, Johannes and Aitken, Andrew P and Bishop, Rob and Rueckert, Daniel and Wang, Zehan},
  booktitle={Proceedings of the IEEE Conference on Computer Vision and Pattern Recognition},
  pages={1874--1883},
  year={2016}
}

@inproceedings{sapa,
 author = {Lu, Hao and Liu, Wenze and Ye, Zixuan and Fu, Hongtao and Liu, Yuliang and Cao, Zhiguo},
 booktitle = {Advances in Neural Information Processing Systems},
 pages = {20889--20901},
 title = {SAPA: Similarity-Aware Point Affiliation for Feature Upsampling},
 year = {2022}
}

@inproceedings{fade,
  title={FADE: Fusing the assets of decoder and encoder for task-agnostic upsampling},
  author={Lu, Hao and Liu, Wenze and Fu, Hongtao and Cao, Zhiguo},
  booktitle={Proceedings of the European Conference on Computer Vision},
  pages={231--247},
  year={2022},
  organization={Springer}
}

@inproceedings{a2u,
  title={Learning affinity-aware upsampling for deep image matting},
  author={Dai, Yutong and Lu, Hao and Shen, Chunhua},
  booktitle={Proceedings of the IEEE/CVF Conference on Computer Vision and Pattern Recognition},
  pages={6841--6850},
  year={2021}
}

@misc{mmseg,
    title={{MMSegmentation}: OpenMMLab Semantic Segmentation Toolbox and Benchmark},
    author={MMSegmentation Contributors},
    howpublished = {\url{https://github.com/open-mmlab/mmsegmentation}},
    year={2020}
}

@inproceedings{segmenter,
  title={Segmenter: Transformer for semantic segmentation},
  author={Strudel, Robin and Garcia, Ricardo and Laptev, Ivan and Schmid, Cordelia},
  booktitle={Proc. IEEE Int. Conf. Comput. Vis. (ICCV)},
  pages={7262--7272},
  year={2021}
}

@inproceedings{psanet,
  title={Psanet: Point-wise spatial attention network for scene parsing},
  author={Zhao, Hengshuang and Zhang, Yi and Liu, Shu and Shi, Jianping and Loy, Chen Change and Lin, Dahua and Jia, Jiaya},
  booktitle={Proceedings of the European Conference on Computer Vision},
  pages={267--283},
  year={2018}
}

@inproceedings{pspnet,
  title={Pyramid scene parsing network},
  author={Zhao, Hengshuang and Shi, Jianping and Qi, Xiaojuan and Wang, Xiaogang and Jia, Jiaya},
  booktitle={Proceedings of the IEEE Conference on Computer Vision and Pattern Recognition},
  pages={2881--2890},
  year={2017}
}

@inproceedings{oneformer,
  title={Oneformer: One transformer to rule universal image segmentation},
  author={Jain, Jitesh and Li, Jiachen and Chiu, Mang Tik and Hassani, Ali and Orlov, Nikita and Shi, Humphrey},
  booktitle={Proceedings of the IEEE/CVF Conference on Computer Vision and Pattern Recognition},
  pages={2989--2998},
  year={2023}
}

@inproceedings{he2017mask,
  title={Mask r-cnn},
  author={He, Kaiming and Gkioxari, Georgia and Doll{\'a}r, Piotr and Girshick, Ross},
  booktitle={Proc. IEEE Int. Conf. Comput. Vis. (ICCV)},
  pages={2961--2969},
  year={2017}
}

@inproceedings{dfine,
title={D-{FINE}: Redefine Regression Task of {DETR}s as Fine-grained Distribution Refinement},
author={Yansong Peng and Hebei Li and Peixi Wu and Yueyi Zhang and Xiaoyan Sun and Feng Wu},
booktitle={The Thirteenth International Conference on Learning Representations},
year={2025},
}

@inproceedings{gflv2,
  title={Generalized focal loss v2: Learning reliable localization quality estimation for dense object detection},
  author={Li, Xiang and Wang, Wenhai and Hu, Xiaolin and Li, Jun and Tang, Jinhui and Yang, Jian},
  booktitle={Proceedings of the IEEE/CVF Conference on Computer Vision and Pattern Recognition},
  pages={11632--11641},
  year={2021}
}

@inproceedings{fcos,
  title={Fcos: Fully convolutional one-stage object detection},
  author={Tian, Zhi and Shen, Chunhua and Chen, Hao and He, Tong},
  booktitle={Proc. IEEE Int. Conf. Comput. Vis. (ICCV)},
  pages={9627--9636},
  year={2019}
}

@inproceedings{reppointsv2,
  title={Reppoints v2: Verification meets regression for object detection},
  author={Chen, Yihong and Zhang, Zheng and Cao, Yue and Wang, Liwei and Lin, Stephen and Hu, Han},
  booktitle={Advances in Neural Information Processing Systems},
  volume={33},
  pages={5621--5631},
  year={2020}
}

@inproceedings{yolov10,
  title={Yolov10: Real-time end-to-end object detection},
  author={Wang, Ao and Chen, Hui and Liu, Lihao and Chen, Kai and Lin, Zijia and Han, Jungong and others},
  booktitle={Advances in Neural Information Processing Systems},
  pages={107984--108011},
  year={2024}
}

@inproceedings{pconv,
  title={Scale-equalizing pyramid convolution for object detection},
  author={Wang, Xinjiang and Zhang, Shilong and Yu, Zhuoran and Feng, Litong and Zhang, Wayne},
  booktitle={Proceedings of the IEEE/CVF Conference on Computer Vision and Pattern Recognition},
  pages={13359--13368},
  year={2020}
}

@inproceedings{rcnet,
  title={RCNet: Reverse feature pyramid and cross-scale shift network for object detection},
  author={Zong, Zhuofan and Cao, Qianggang and Leng, Biao},
  booktitle={Proceedings of the 29th ACM International Conference on Multimedia},
  pages={5637--5645},
  year={2021}
}

@inproceedings{segformer,
  title={SegFormer: Simple and efficient design for semantic segmentation with transformers},
  author={Xie, Enze and Wang, Wenhai and Yu, Zhiding and Anandkumar, Anima and Alvarez, Jose M and Luo, Ping},
  booktitle={Advances in Neural Information Processing Systems},
  volume={34},
  pages={12077--12090},
  year={2021}
}

@inproceedings{segnext,
  title={Segnext: Rethinking convolutional attention design for semantic segmentation},
  author={Guo, Meng-Hao and Lu, Cheng-Ze and Hou, Qibin and Liu, Zhengning and Cheng, Ming-Ming and Hu, Shi-Min},
  booktitle={Advances in Neural Information Processing Systems},
  volume={35},
  pages={1140--1156},
  year={2022}
}

@inproceedings{goldyolo,
  title={Gold-YOLO: Efficient object detector via gather-and-distribute mechanism},
  author={Wang, Chengcheng and He, Wei and Nie, Ying and Guo, Jianyuan and Liu, Chuanjian and Wang, Yunhe and Han, Kai},
  booktitle={Advances in Neural Information Processing Systems},
  volume={36},
  pages={51094--51112},
  year={2023}
}

@inproceedings{fpt,
  title={Feature pyramid transformer},
  author={Zhang, Dong and Zhang, Hanwang and Tang, Jinhui and Wang, Meng and Hua, Xiansheng and Sun, Qianru},
  booktitle={Proceedings of the European Conference on Computer Vision},
  pages={323--339},
  year={2020},
  organization={Springer}
}

@inproceedings{ablation_cam,
  title={Ablation-cam: Visual explanations for deep convolutional network via gradient-free localization},
  author={Ramaswamy, Harish Guruprasad and others},
  booktitle={Proceedings of the IEEE/CVF Winter Conference on Applications of Computer Vision},
  pages={983--991},
  year={2020}
}

@inproceedings{resnet,
  title={Deep residual learning for image recognition},
  author={He, Kaiming and Zhang, Xiangyu and Ren, Shaoqing and Sun, Jian},
  booktitle={Proceedings of the IEEE Conference on Computer Vision and Pattern Recognition},
  pages={770--778},
  year={2016}
}

@article{pascalvoc,
  title={The pascal visual object classes (voc) challenge},
  author={Everingham, Mark and Van Gool, Luc and Williams, Christopher KI and Winn, John and Zisserman, Andrew},
  journal={International Journal of Computer Vision},
  volume={88},
  pages={303--338},
  year={2010},
  publisher={Springer}
}

@ARTICLE{carafe++,
  author={Wang, Jiaqi and Chen, Kai and Xu, Rui and Liu, Ziwei and Loy, Chen Change and Lin, Dahua},
  journal={IEEE Transactions on Pattern Analysis and Machine Intelligence}, 
  title={CARAFE++: Unified Content-Aware ReAssembly of FEatures}, 
  year={2022},
  volume={44},
  number={9},
  pages={4674-4687},
}

@inproceedings{dysample,
  title={Learning to upsample by learning to sample},
  author={Liu, Wenze and Lu, Hao and Fu, Hongtao and Cao, Zhiguo},
  booktitle={Proc. IEEE Int. Conf. Comput. Vis. (ICCV)},
  pages={6027--6037},
  year={2023}
}

@article{frequency_aware,
  title={Frequency-aware feature fusion for dense image prediction},
  author={Chen, Linwei and Fu, Ying and Gu, Lin and Yan, Chenggang and Harada, Tatsuya and Huang, Gao},
  journal={IEEE Transactions on Pattern Analysis and Machine Intelligence},
  year={2024},
  publisher={IEEE}
}

@inproceedings{dynamask,
  title={DynaMask: dynamic mask selection for instance segmentation},
  author={Li, Ruihuang and He, Chenhang and Li, Shuai and Zhang, Yabin and Zhang, Lei},
  booktitle={Proceedings of the IEEE/CVF Conference on Computer Vision and Pattern Recognition},
  pages={11279--11288},
  year={2023}
}

@inproceedings{setr,
  title={Rethinking semantic segmentation from a sequence-to-sequence perspective with transformers},
  author={Zheng, Sixiao and Lu, Jiachen and Zhao, Hengshuang and Zhu, Xiatian and Luo, Zekun and Wang, Yabiao and Fu, Yanwei and Feng, Jianfeng and Xiang, Tao and Torr, Philip HS and others},
  booktitle={Proceedings of the IEEE/CVF Conference on Computer Vision and Pattern Recognition},
  pages={6881--6890},
  year={2021}
}

@inproceedings{faster_rcnn,
author = {Ren, Shaoqing and He, Kaiming and Girshick, Ross and Sun, Jian},
title = {Faster R-CNN: towards real-time object detection with region proposal networks},
year = {2015},
booktitle = {Advances in Neural Information Processing Systems},
pages = {91–99},
}

@inproceedings{cityscapes,
  title={The cityscapes dataset for semantic urban scene understanding},
  author={Cordts, Marius and Omran, Mohamed and Ramos, Sebastian and Rehfeld, Timo and Enzweiler, Markus and Benenson, Rodrigo and Franke, Uwe and Roth, Stefan and Schiele, Bernt},
  booktitle={Proceedings of the IEEE Conference on Computer Vision and Pattern Recognition},
  pages={3213--3223},
  year={2016}
}

@inproceedings{upernet,
  title={Unified perceptual parsing for scene understanding},
  author={Xiao, Tete and Liu, Yingcheng and Zhou, Bolei and Jiang, Yuning and Sun, Jian},
  booktitle={Proceedings of the European Conference on Computer Vision},
  pages={418--434},
  year={2018}
}

@inproceedings{rtdetrv1,
  title={Detrs beat yolos on real-time object detection},
  author={Zhao, Yian and Lv, Wenyu and Xu, Shangliang and Wei, Jinman and Wang, Guanzhong and Dang, Qingqing and Liu, Yi and Chen, Jie},
  booktitle={Proceedings of the IEEE/CVF Conference on Computer Vision and Pattern Recognition},
  pages={16965--16974},
  year={2024}
}

@inproceedings{mask_dino,
  title={Mask dino: Towards a unified transformer-based framework for object detection and segmentation},
  author={Li, Feng and Zhang, Hao and Xu, Huaizhe and Liu, Shilong and Zhang, Lei and Ni, Lionel M and Shum, Heung-Yeung},
  booktitle={Proceedings of the IEEE/CVF Conference on Computer Vision and Pattern Recognition},
  pages={3041--3050},
  year={2023}
}

@inproceedings{fpn,
  title={Feature pyramid networks for object detection},
  author={Lin, Tsung-Yi and Doll{\'a}r, Piotr and Girshick, Ross and He, Kaiming and Hariharan, Bharath and Belongie, Serge},
  booktitle={Proceedings of the IEEE Conference on Computer Vision and Pattern Recognition},
  pages={2117--2125},
  year={2017}
}

@inproceedings{fcn_semantic,
  title={Fully convolutional networks for semantic segmentation},
  author={Long, Jonathan and Shelhamer, Evan and Darrell, Trevor},
  booktitle={Proceedings of the IEEE Conference on Computer Vision and Pattern Recognition},
  pages={3431--3440},
  year={2015}
}

@INPROCEEDINGS{mask2former,
  author={Cheng, Bowen and Misra, Ishan and Schwing, Alexander G. and Kirillov, Alexander and Girdhar, Rohit},
  booktitle={Proceedings of the IEEE/CVF Conference on Computer Vision and Pattern Recognition}, 
  title={Masked-attention Mask Transformer for Universal Image Segmentation}, 
  year={2022},
  volume={},
  number={},
  pages={1280-1289},
}

@inproceedings{efficientdet,
  title={Efficientdet: Scalable and efficient object detection},
  author={Tan, Mingxing and Pang, Ruoming and Le, Quoc V},
  booktitle={Proceedings of the IEEE/CVF Conference on Computer Vision and Pattern Recognition},
  pages={10781--10790},
  year={2020}
}

@ARTICLE{HRNEtforvisual,
  author={Wang, Jingdong and Sun, Ke and Cheng, Tianheng and Jiang, Borui and Deng, Chaorui and Zhao, Yang and Liu, Dong and Mu, Yadong and Tan, Mingkui and Wang, Xinggang and Liu, Wenyu and Xiao, Bin},
  journal={IEEE Transactions on Pattern Analysis and Machine Intelligence}, 
  title={Deep High-Resolution Representation Learning for Visual Recognition}, 
  year={2021},
  volume={43},
  number={10},
  pages={3349-3364},
}

@inproceedings{fapn,
  title={FaPN: Feature-aligned pyramid network for dense image prediction},
  author={Huang, Shihua and Lu, Zhichao and Cheng, Ran and He, Cheng},
  booktitle={Proc. IEEE Int. Conf. Comput. Vis. (ICCV)},
  pages={864--873},
  year={2021}
}

@inproceedings{cai2018cascade,
  title={Cascade r-cnn: Delving into high quality object detection},
  author={Cai, Zhaowei and Vasconcelos, Nuno},
  booktitle={Proceedings of the IEEE Conference on Computer Vision and Pattern Recognition},
  pages={6154--6162},
  year={2018}
}

@inproceedings{dcnv4,
  title={Efficient deformable convnets: Rethinking dynamic and sparse operator for vision applications},
  author={Xiong, Yuwen and Li, Zhiqi and Chen, Yuntao and Wang, Feng and Zhu, Xizhou and Luo, Jiapeng and Wang, Wenhai and Lu, Tong and Li, Hongsheng and Qiao, Yu and others},
  booktitle={Proceedings of the IEEE/CVF Conference on Computer Vision and Pattern Recognition},
  pages={5652--5661},
  year={2024}
}

@inproceedings{deformable_detr,
    title={Deformable DETR: Deformable Transformers for End-to-End Object Detection},
    author={Xizhou Zhu and Weijie Su and Lewei Lu and Bin Li and Xiaogang Wang and Jifeng Dai},
    booktitle={International Conference on Learning Representations},
    year={2021},
}

@inproceedings{detr,
  title={End-to-end object detection with transformers},
  author={Carion, Nicolas and Massa, Francisco and Synnaeve, Gabriel and Usunier, Nicolas and Kirillov, Alexander and Zagoruyko, Sergey},
  booktitle={Proceedings of the European Conference on Computer Vision},
  pages={213--229},
  year={2020},
  organization={Springer}
}

@inproceedings{pointrend,
  title={Pointrend: Image segmentation as rendering},
  author={Kirillov, Alexander and Wu, Yuxin and He, Kaiming and Girshick, Ross},
  booktitle={Proceedings of the IEEE/CVF Conference on Computer Vision and Pattern Recognition},
  pages={9799--9808},
  year={2020}
}

@inproceedings{gn,
  title={Group normalization},
  author={Wu, Yuxin and He, Kaiming},
  booktitle={Proceedings of the European Conference on Computer Vision},
  pages={3--19},
  year={2018}
}

@inproceedings{nas-fpn,
  title={Nas-fpn: Learning scalable feature pyramid architecture for object detection},
  author={Ghiasi, Golnaz and Lin, Tsung-Yi and Le, Quoc V},
  booktitle={Proceedings of the IEEE/CVF Conference on Computer Vision and Pattern Recognition},
  pages={7036--7045},
  year={2019}
}

@inproceedings{zhao2021graphfpn,
  title={GraphFPN: Graph feature pyramid network for object detection},
  author={Zhao, Gangming and Ge, Weifeng and Yu, Yizhou},
  booktitle={Proc. IEEE Int. Conf. Comput. Vis. (ICCV)},
  pages={2763--2772},
  year={2021}
}

@inproceedings{liu2018path,
  title={Path aggregation network for instance segmentation},
  author={Liu, Shu and Qi, Lu and Qin, Haifang and Shi, Jianping and Jia, Jiaya},
  booktitle={Proceedings of the IEEE Conference on Computer Vision and Pattern Recognition},
  pages={8759--8768},
  year={2018}
}

@inproceedings{wang2019carafe,
  title={Carafe: Content-aware reassembly of features},
  author={Wang, Jiaqi and Chen, Kai and Xu, Rui and Liu, Ziwei and Loy, Chen Change and Lin, Dahua},
  booktitle={Proc. IEEE Int. Conf. Comput. Vis. (ICCV)},
  pages={3007--3016},
  year={2019}
}

@inproceedings{a2-fpn,
  title={A2-FPN: Attention aggregation based feature pyramid network for instance segmentation},
  author={Hu, Miao and Li, Yali and Fang, Lu and Wang, Shengjin},
  booktitle={Proceedings of the IEEE/CVF Conference on Computer Vision and Pattern Recognition},
  pages={15343--15352},
  year={2021}
}

@inproceedings{coco,
  title={Microsoft coco: Common objects in context},
  author={Lin, Tsung-Yi and Maire, Michael and Belongie, Serge and Hays, James and Perona, Pietro and Ramanan, Deva and Doll{\'a}r, Piotr and Zitnick, C Lawrence},
  booktitle={Proceedings of the European Conference on Computer Vision},
  pages={740--755},
  year={2014},
  organization={Springer}
}

@inproceedings{augfpn,
  title={Augfpn: Improving multi-scale feature learning for object detection},
  author={Guo, Chaoxu and Fan, Bin and Zhang, Qian and Xiang, Shiming and Pan, Chunhong},
  booktitle={Proceedings of the IEEE/CVF Conference on Computer Vision and Pattern Recognition},
  pages={12595--12604},
  year={2020}
}

@article{mmdet,
  title={MMDetection: Open mmlab detection toolbox and benchmark},
  author={Chen, Kai and Wang, Jiaqi and Pang, Jiangmiao and Cao, Yuhang and Xiong, Yu and Li, Xiaoxiao and Sun, Shuyang and Feng, Wansen and Liu, Ziwei and Xu, Jiarui and others},
  journal={arXiv preprint arXiv:1906.07155},
  year={2019}
}

@inproceedings{focal_loss,
  title={Focal loss for dense object detection},
  author={Lin, Tsung-Yi and Goyal, Priya and Girshick, Ross and He, Kaiming and Doll{\'a}r, Piotr},
  booktitle={Proc. IEEE Int. Conf. Comput. Vis. (ICCV)},
  pages={2980--2988},
  year={2017}
}
	
	\appendix
	\newpage
	\section{Algorithm Procedures of \texorpdfstring{$A^3$-FPN}{A3-FPN}}
	\begin{algorithm}[!ht]
		\caption{Bottom-up Asymptotic Content-Aware Pyramid Attention Network}
		\label{alg:a3fpn}
		
		\textbf{Input}: $n$ hierarchical features $\{X_{1}, X_{2},\ldots, X_{n}\}$ from the backbone, which will go through $\textbf{m}$ column transformations in $A^3$-FPN.
		
		Initialize $\{Z_1, Z_2,\ldots, Z_n\}$
		
		\For{$j \gets 1$ to $\mathbf{m}$}{
			$\textbf{min}=\min(j+1, n)$
			
			\For{$i \gets 1$ to $\mathbf{min}$}{
				
				\For{$k \gets 1$ to $\mathbf{min}$}{
					{$k > i$}:
					Upsample $X_{k}$ to $X_{k}^{samp}$
					
					{$k < i$}:
					Downsample $X_{k}$ to $X_{k}^{samp}$
				}
				
				\[
				\left\{X_1^{offset}, \cdots, X_{i-1}^{offset}, X_{i+1}^{offset}, \cdots, X_{\textbf{min}}^{offset} \right\} \gets \mathbf{Offset Gererator}(\left\{X_1^{samp}, \cdots, X_i, \cdots, X_{\textbf{min}}^{samp}\right\})
				\]
				
				\For{$k \gets 1$ to $\mathbf{min}$}{
					{$k \neq i$}: 
					$X^{rs}_k \gets \mathbf{Resampler}(X_{k}^{samp}, X_k^{offset})$
				}
				
				\[
				\left\{W_1, \cdots, W_i,\cdots, W_{\textbf{min}} \right\} \gets \mathbf{Context Weight Gererator} (\left\{X_1^{rs}, \cdots, X_i, \cdots, X_{\textbf{min}}^{rs} \right\})
				\]
				
				Feature fusion: \[Y_i=(\sum_{n=1, n \neq i}^{\textbf{min}}{W_n}\bigotimes X^{rs}_{n}) + W_i \bigotimes X_{i}\]
				
				Compute spatial weights:
				\[
				Y_i^{\mathrm{std}}=GN(Y_i)=\alpha\frac{Y_i-\mu}{\sqrt{\sigma^2}}+\beta, 
				\omega_i=\left\{\frac{\alpha_i}{\sum_{j=1}^{c_i} \alpha_j}, i=1,2, \cdots,c_i\right\}
				\]
				
				Attain information weights: 
				\[\omega_i^1 = 
				\begin{cases}
					\text{Sigmoid}(\omega_i), & \text{Sigmoid}(\omega_i) \leq \text{0.5}; \\
					1, & \text{Sigmoid}(\omega_i) > \text{0.5},
				\end{cases},  
				\omega_i^2 = 
				\begin{cases} 
					\text{Sigmoid}(\omega_i), & \text{Sigmoid}(\omega_i) \geq \text{0.5}; \\
					0, & \text{Sigmoid}(\omega_i) < \text{0.5}.
				\end{cases}
				\]
				
				Feature reassembly:  \[Z_i =\bigcup_{c=1}^{C} \left( Y_i^c \times \omega_i^{1c} + Y_i^{C-c+1} \times \omega_i^{2c} \right) \]}
			
			\[\left\{X_1, \cdots, X_{\textbf{min}},X_{\textbf{min}+1},\cdots, X_n \right\} \gets \left\{Z_1, \cdots, Z_{\textbf{min}}, X_{\textbf{min}+1}, \cdots, X_n\right\}\]
		}
		\textbf{Output}: $n$ refined multi-scale features maps $\{Z_1, Z_2,\ldots, Z_n\}$
	\end{algorithm}
	
	\newpage
	\section{Top-down Asymptotically Disentangled Framework}\label{top_down}
	\begin{figure}[htbp]
		\centering
		\includegraphics[width=0.75\linewidth]{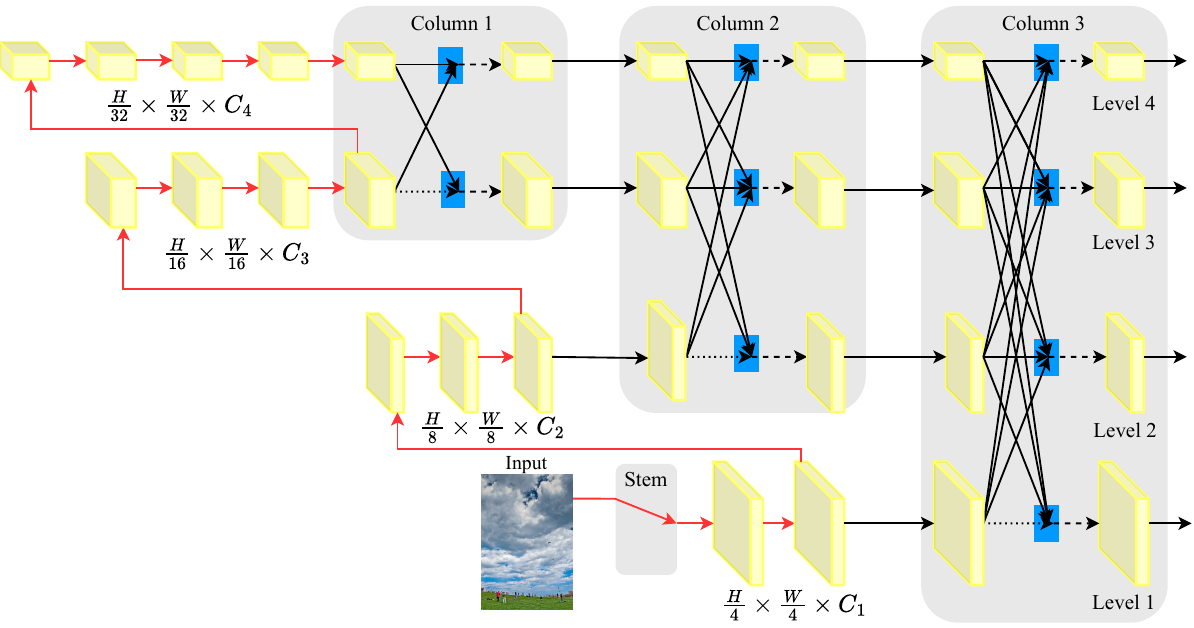}
		\caption{Illustration of A3-FPN with the top-down asymptotic disentangled framework.}
		\label{fig:top_down_a3fpn}
	\end{figure}
	
	\section{Process Visualization in MCAtten}\label{vis_generator}
	\begin{figure}[htbp]
		\centering
		\includegraphics[width=0.7\linewidth]{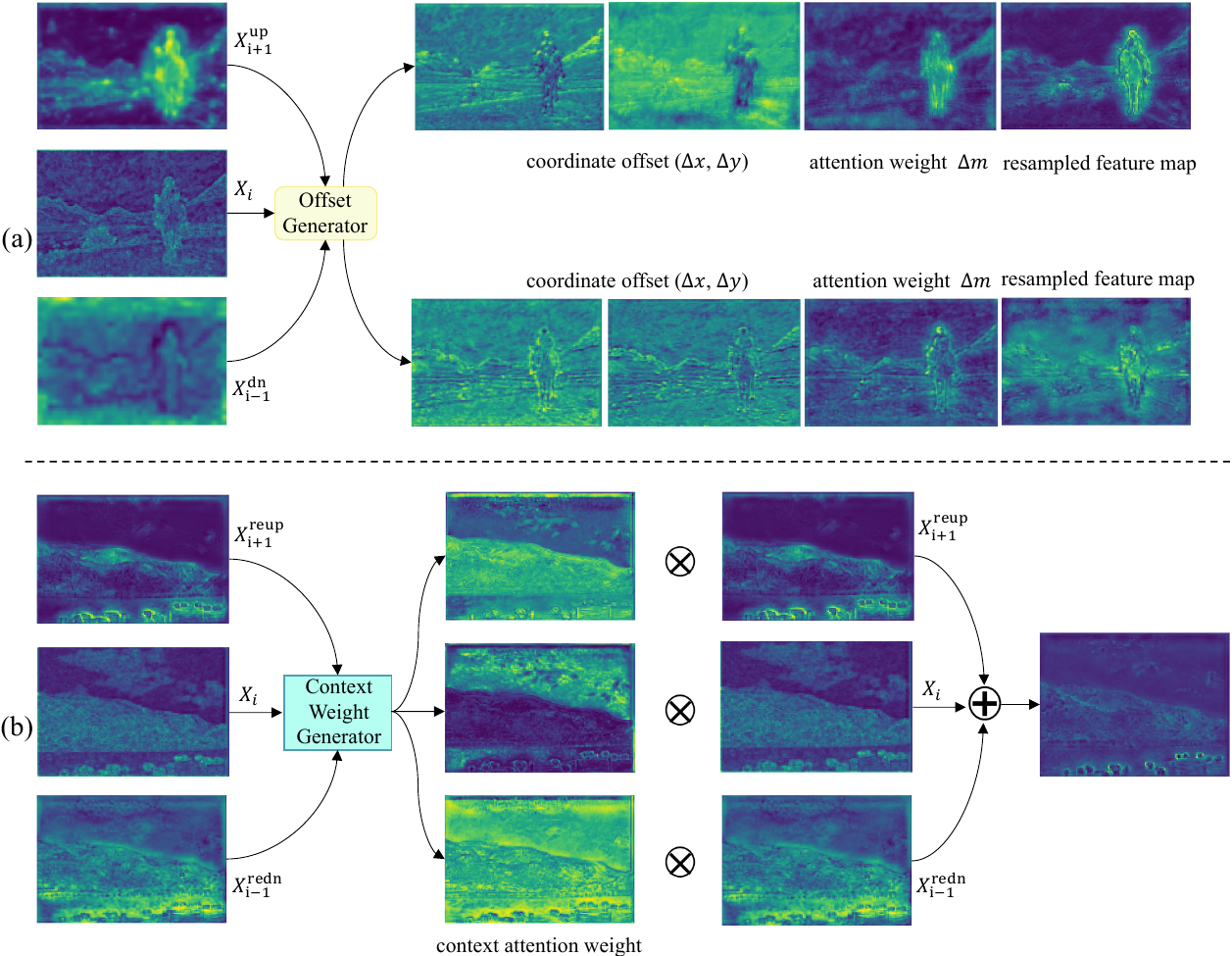}
		\caption{(a) Visualization of coordinate offsets and the corresponding attention weight in the offset generator. (b) Fusion process visualization in MCAtten.}
		\label{fig:vis_generator}
	\end{figure}
	
	\newpage
	\section{More qualitative evaluations of \texorpdfstring{$A^3$-FPN}{A3-FPN} on the instance segmentation task}
	\begin{figure}[htbp]
		\centering
		\includegraphics[width=0.64\linewidth]{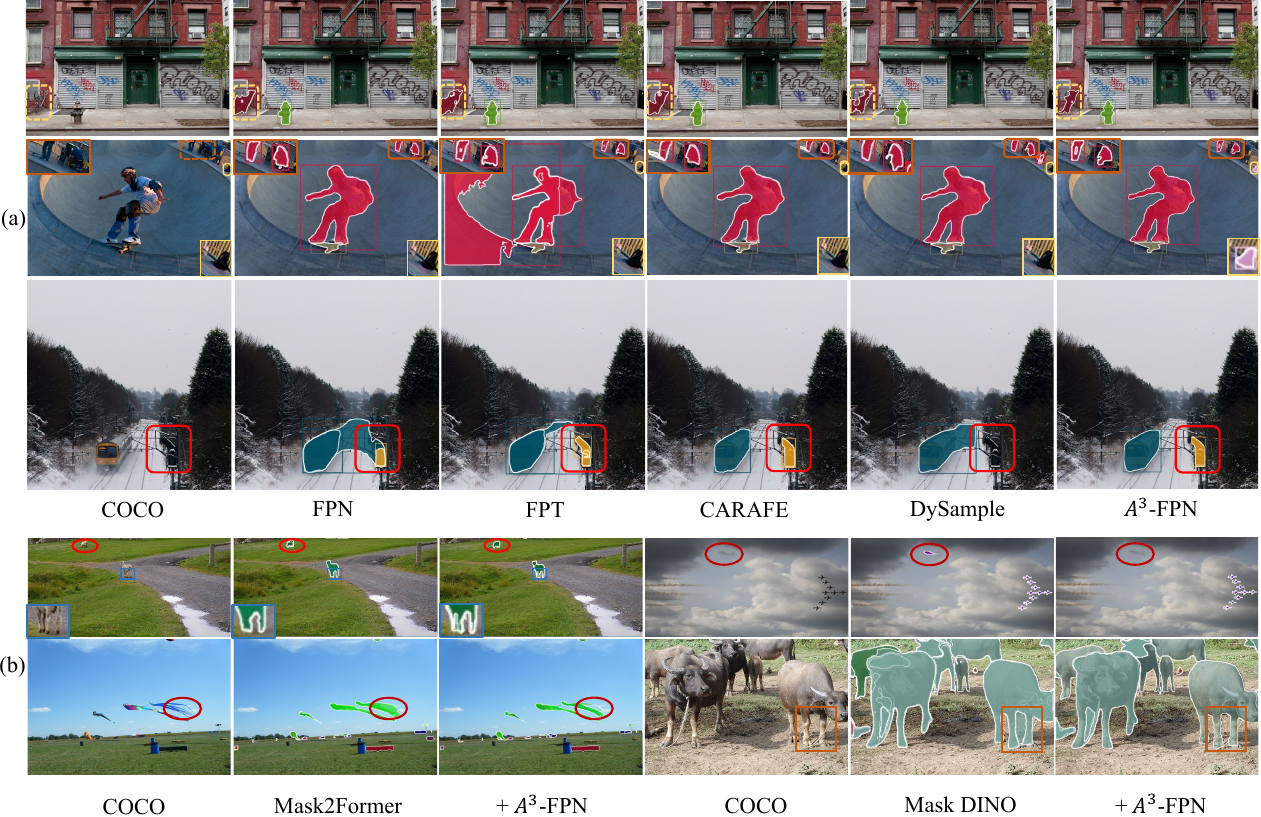}
		\caption{(a) Instance segmentation results of Mask RCNN\cite{he2017mask} with different feature fusion approaches, including FPN\cite{fpn}, FPT\cite{fpt}, DySample\cite{dysample}, CARAFE\cite{wang2019carafe} and our $A^3$-FPN. (b) Qualitative comparison between some unified transformer-based models (Mask2Former\cite{mask2former} and Mask DINO\cite{mask_dino}) with and without integrating $A^3$-FPN on the instance segmentation task.}
		\label{fig:instance_seg}
	\end{figure}
	
	\section{More qualitative evaluations of \texorpdfstring{$A^3$-FPN}{A3-FPN} on the semantic segmentation task}
	\begin{figure}[!htbp]
		\centering
		\includegraphics[width=0.62\linewidth]{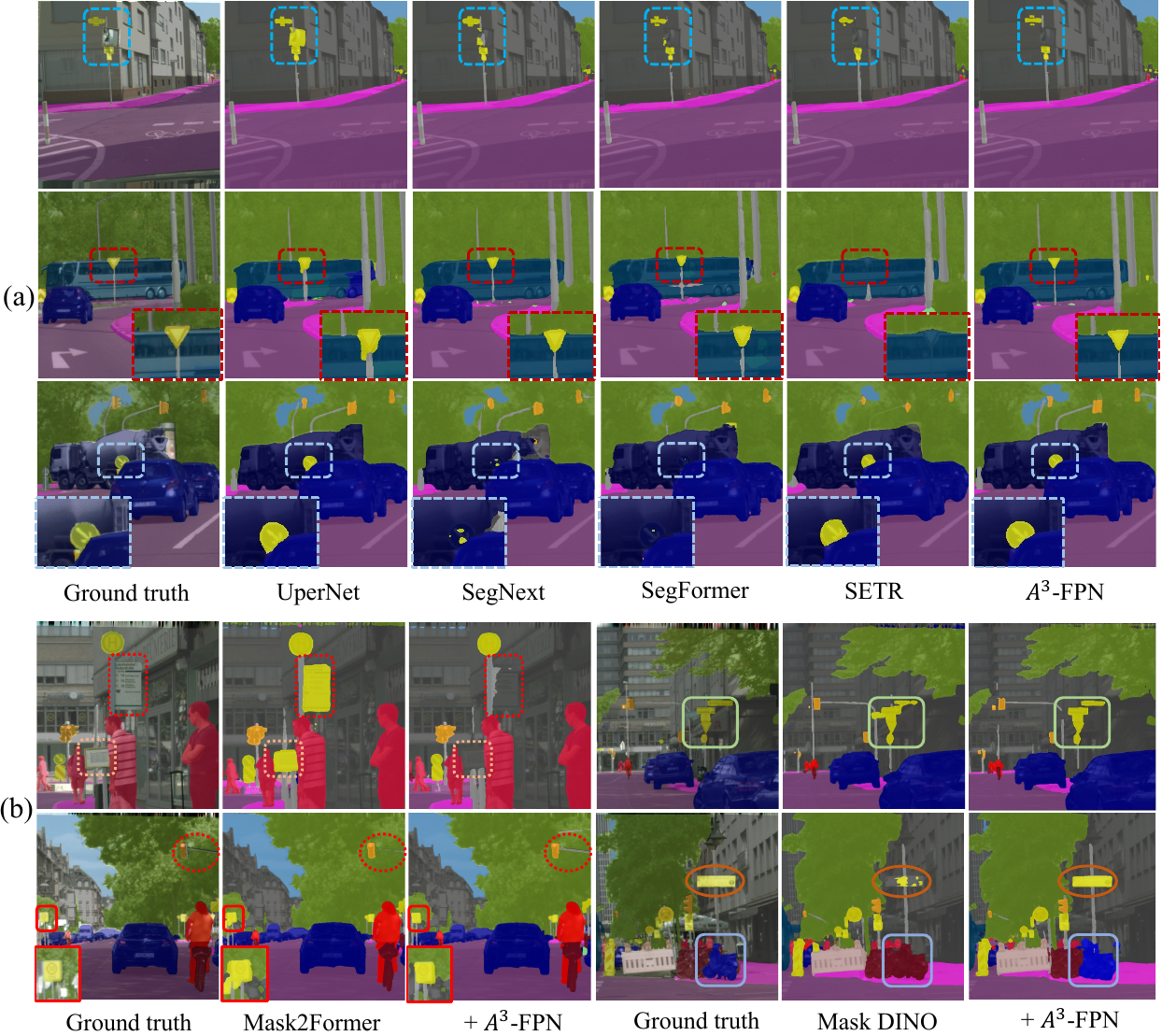}
		\caption{(a) Visualization on Cityscapes validation set using different semantic segmentors, which are UperNet\cite{upernet}, SegNext\cite{segnext}, SegFormer\cite{segformer}, SETR\cite{setr}, and our $A^3$-FPN. (b) Qualitative evaluation of unified transformer-based models (Mask2Former\cite{mask2former} and Mask DINO\cite{mask_dino}) with and without integrating $A^3$-FPN on the semantic segmentation task.}
		\label{fig:sem_seg}
	\end{figure}
	
	\newpage
	\section{Hyperparameter Setting between \texorpdfstring{$A^3$-FPN}{A3-FPN} and \texorpdfstring{$A^3$-FPN-Lite}{A3-FPN-Lite}}
	\label{hyerparameter}
	\begin{table}[htbp]
		\centering
		\caption{Hyperparameters setting between $A^3$-FPN and $A^3$-FPN-Lite.}
		\renewcommand{\arraystretch}{1.0}
		\label{tab:hyperparameter}
		\resizebox{0.5\linewidth}{!}{
			\begin{tabular}{l|c|c}
				\toprule
				Hyperparameter  & $A^3$-FPN & $A^3$-FPN-Lite \\
				\midrule
				Squeeze & [1, 2, 4, 4] & [1, 2, 4, 8] \\
				Using Resampling & [True, True, True] & [False, False, True] \\
				Compress Channels & [16, 16, 16, 32] & [16, 16, 16, 16] \\
				GN Group & [16, 16, 16, 32] & [16, 16, 16, 16] \\
				RepBlock Number & 2 & 1 \\
				Expansion & 4.0 & 2.0 \\
				Resample Group & [16, 16, 16, 32] & [16, 16, 16, 16] \\
				Offset Scale & 2.0 & 1.0 \\
				Norm after Resampling & LN & LN \\
				Output Bias in Resampler & True & True \\
				DWConv Kernel & 3 & 3 \\
				\bottomrule
			\end{tabular}
		}
	\end{table}
	
\end{document}